\documentclass[journal]{IEEEtran}
\usepackage{times}
\usepackage{multicol}
\usepackage[bookmarks=true]{hyperref}

\usepackage{xparse}
\usepackage{url}
\usepackage{graphicx}
\usepackage{amsmath}
\usepackage{amssymb}

\usepackage{array}
\usepackage{longtable}
\usepackage{multirow}
\usepackage{float}
\usepackage{mathrsfs}
\usepackage{subcaption}
\usepackage{listings}
\usepackage{rotating}
\usepackage{overpic}

\usepackage[ruled,vlined]{algorithm2e}
\usepackage{etoolbox}%
\providetoggle{beginFlag}
\settoggle{beginFlag}{true}

\usepackage{cleveref}
\crefformat{equation}{(#2#1#3)}
\Crefname{figure}{Fig.}{Figs.}
\crefname{figure}{Fig.}{Figs.}
\crefformat{figure}{#2Fig.~#1#3}
\Crefformat{figure}{#2Fig.~#1#3}
\crefformat{table}{#2Tab.~#1#3}
\Crefformat{table}{#2Tab.~#1#3}
\crefformat{section}{#2Sec.~#1#3}
\Crefformat{section}{#2Sec.~#1#3}
\crefformat{problem}{#2Problem~#1#3}
\Crefformat{problem}{#2Problem~#1#3}
\crefformat{algorithm}{#2Algorithm~#1#3}
\Crefformat{algorithm}{#2Algorithm~#1#3}
\crefname{appsec}{Appendix}{Appendices}

\usepackage{booktabs}

\usepackage{rigidnotation}
\SetConciseNotation
\usepackage{xparse}
\usepackage{tikz}

\newcommand{\lrss}[5]{%
	\setbox1=\hbox{\ensuremath{^{#1}}}%
	\setbox2=\hbox{\ensuremath{_{#2}}}%
	\setbox5=\hbox{\ensuremath{#5}}%
	\setbox6=\hbox{\ensuremath{^{#1#3}}}%
	\setbox7=\hbox{\ensuremath{_{#2#4}}}%
	\setbox8=\hbox{\ensuremath{^{#3}}}%
	\setbox9=\hbox{\ensuremath{_{#4}}}%
	\hspace{\ifnum\wd1>\wd2\wd1\else\wd2\fi}%
	\ensuremath{\copy5%
		^{\hspace{-\wd1}\hspace{\wd1}\hspace{\wd8}%
			\hspace{-\wd6}\hspace{-\wd5}#1\hspace{\wd5}#3}%
		_{\hspace{-\wd2}\hspace{\wd2}\hspace{\wd9}%
			\hspace{-\wd7}\hspace{-\wd5}#2\hspace{\wd5}#4}%
}}

\NewDocumentCommand\bbm{}{ \begin{bmatrix} }
\NewDocumentCommand\ebm{}{ \end{bmatrix} }
\NewDocumentCommand\Vector{m}{ \boldsymbol{\mathbf{#1}} }
\NewDocumentCommand\Matrix{m}{ \boldsymbol{\mathbf{#1}} }
\NewDocumentCommand\UnitVec{m}{ \Vector{\hat{#1}} }

\NewDocumentCommand\Norm{m}{\left\Vert#1\right\Vert }

\NewDocumentCommand\Transpose{}{^\mathsf{T}}

\NewDocumentCommand\Real{}{ \mathbb{R} }

\NewDocumentCommand\LieGroupSE{m}{ \mathrm{SE}(#1) }

\NewDocumentCommand\IdentityMatrix{m}{ \Matrix{\IdentitySymbol}_{#1} }

\NewDocumentCommand\CoordinateFrame{m}{ \underrightarrow{\Matrix{\mathcal{F}}}_{#1} }

\NewDocumentCommand\IdentitySymbol{}{ 1 }
\NewDocumentCommand\Skew{m}{\left[#1\right]_\times}

\NewDocumentCommand\DotP{}{ \cdot }
\NewDocumentCommand\CrossP{}{ \times }

\NewDocumentCommand\Mass{}{ m }

\NewDocumentCommand\Gravity{}{ \Vector{g} }
\NewDocumentCommand\Force{}{ \Vector{f} }

\NewDocumentCommand\AngularVelocity{}{ \Vector{\omega} }

\NewDocumentCommand\AAxis{}{ \UnitVec{\AngularVelocity} }
\NewDocumentCommand\AAngle{}{ \theta }

\RenewDocumentCommand\CoordinateFrame{m}{ \Matrix{\mathcal{F}}_{#1} }
\NewDocumentCommand\CFrame{m}{ \CoordinateFrame{#1} }
\NewDocumentCommand\CSys{m}{ \{#1\} }

\NewRigidNotation{Normal}{n}
\NewRigidNotation{Edge}{g}
\NewRigidNotation{Side}{s}
\NewRigidNotation{VVec}{v}
\NewRigidNotation{UVec}{u}
\NewRigidNotation{WVec}{w}
\NewRigidNotation{Axis}{a}
\NewRigidNotation{EForce}{e}
\NewRigidNotation{RForce}{f}
\NewRigidNotation{Wrench}{w}
\NewRigidNotation{TVec}{\tau}
\NewRigidNotation{Disp}{d}
\NewRigidNotation{Twist}{\nu}

\def\Mag{{s}}

\newcommand{\rev}[1]{\textcolor{black}{#1}}
\newcommand{\revv}[1]{\textcolor{black}{#1}}

\IEEEtriggercmd{\newpage}
\IEEEtriggeratref{44}

\begin{document}

\title{Stable Object Placement Planning\\ from Contact Point Robustness}

\author{Philippe Nadeau and Jonathan Kelly$^\ddagger$
	\thanks{All authors are with the STARS Laboratory at the University of Toronto Institute for Aerospace Studies, Toronto, Ontario, Canada. {\tt\footnotesize <firstname>.<lastname>@robotics.utias.utoronto.ca}}
	\thanks{$^\ddagger$Jonathan Kelly is a Vector Institute Faculty Affiliate. This research was supported in part by the Canada Research Chairs program.}}

\maketitle

\begin{abstract}
We introduce a planner designed to guide robot manipulators in stably placing objects within complex scenes.
Our proposed method reverses the traditional approach to object placement: our planner selects contact points first and then determines a placement pose that solicits the selected points.
This is instead of sampling poses, identifying contact points, and evaluating pose quality.
Our algorithm facilitates stability-aware object placement planning, imposing no restrictions on object shape, convexity, or mass density homogeneity, while avoiding combinatorial computational complexity.
Our proposed stability heuristic enables our planner to find a solution about 20 times faster when compared to the same algorithm not making use of the heuristic and eight times faster than a state-of-the-art method that uses the traditional sample-and-evaluate approach.
The proposed planner is also more successful in finding stable placements than the five other benchmarked algorithms.
Derived from first principles and validated in ten real robot experiments, our approach provides a general and scalable solution to the problem of rigid object placement planning.
\end{abstract}


\IEEEpeerreviewmaketitle

\section{Introduction}
\begin{figure}[t]
    \centering
    \vspace{1mm}
    \includegraphics[width=1\linewidth]{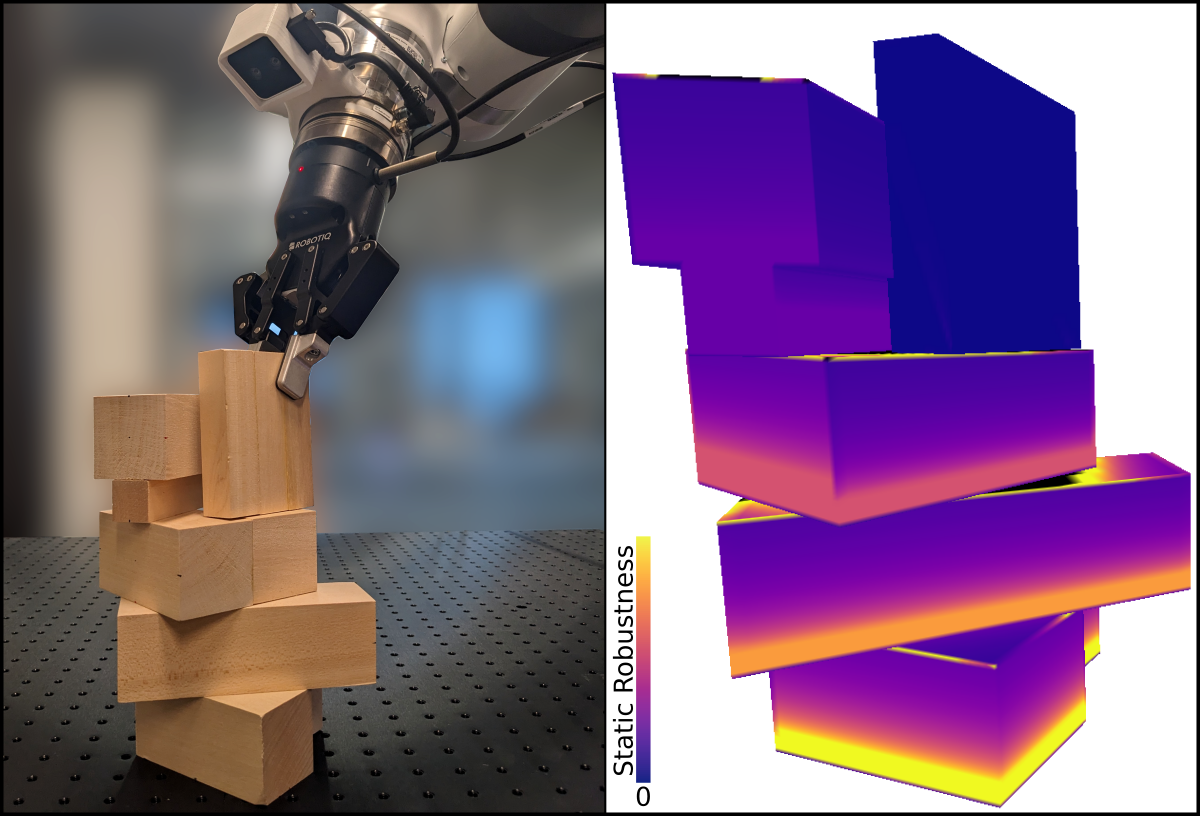}
    \caption{Intricate, stable structure generated by our object placement planner. The surface is shaded according to robustness to perturbation by external forces; we use this measure to inform our proposed planner.}
    \label{fig:frontpage}
    \vspace{-3mm}
\end{figure}

\PARstart{R}{obots} with the capability to stably place objects in contact with one another hold the potential to reduce the need for human intervention in various tasks, spanning home chores, industrial operations, and work in outdoor environments \cite{alterovitz_robot_2016}.
In domestic settings, tasks like tidying a living space by rearranging objects or organizing garage storage could be automated \cite{srinivas_busboy_2023}.
Effective planning for stable object placement could facilitate safe palletizing of mixed products or enable the autonomous loading of trucks in industrial settings.
In outdoor environments, autonomous excavators could be used in disaster relief through debris removal \cite{wermelinger_grasping_2021}.
However, stably packing or rearranging rigid objects in contact is challenging due to the subtle yet influential force interactions between the objects that determine their stability.
Since the problem of determining force interactions \revv{between rigid objects in frictional contact} is known to be NP-hard \cite{baraff_coping_1991}, many approaches resort to planning heuristics based on shape information only and do not generalize well across tasks.
\rev{Moreover, while existing methods have typically considered geometry and dynamics in two separate steps \cite{chen_planning_2021,haustein_object_2019} we propose to (i) combine them through a \textit{static robustness map}, as shown in \cref{fig:frontpage}, and (ii) define a planner that leverages this map to generate stable placements more efficiently---our two main contributions.}

\rev{A general approach to placement planning requires considering the potential motion of objects under forceful interactions.}
In static assemblies, the mass and centre of mass (henceforth referred to as the inertial parameters), as well as the friction coefficients, are sufficient to characterize an object's resistance to acceleration.
A planner that operates based on these parameters is called \textit{inertia-aware} in this work, with knowledge of the second mass moments not being required in static scenarios.
In sum, this paper introduces an inertia-aware object placement planning algorithm, leveraging a physically-grounded heuristic to generate stable placements \rev{of known objects}, with scalability to tackle large-scale problems.

The remainder of this paper is organized as follows.
\cref{sec:related_work} reviews the literature on assembly stability and object placement planning, highlighting the novelty of our proposed algorithm.
In \cref{sec:inertia_aware_object_placement_planning}, the inertia-aware object placement planning problem is formally defined, while \cref{sec:static_robustness_assessment} introduces a physically-grounded heuristic to assess an assembly's capacity to withstand external forces.
Our proposed planner is detailed in \cref{sec:placement_planning_from_static_robustness}, and the results of more than 1,500 simulation experiments involving six scenes and six algorithms are presented in \cref{sec:experiments}, demonstrating the benefits of our method.
In \cref{sec:real_robot_experiments}, practical validation of our proposed algorithm is carried out through 10 real-world experiments involving 50 placements, in which a robot manipulator builds assemblies while recording any failures that occur.
Finally, \cref{sec:discussion} concludes with insights gained from our simulation and real robot experiments.

\section{Related Work}
\label{sec:related_work}
Early methods to produce stable orientations of an assembly under gravity, for fixturing purposes, include \cite{mattikalli_stability_1993}, which ignores friction.
\rev{In \cite{mattikalli_finding_1996}, friction is taken into account but stability is not guaranteed due to indeterminacies in the distribution of contact forces \cite{pang_stability_2000}, a common issue.}
Selecting a set of feasible contact forces can be done by resorting to optimization-based methods \cite{whiting_structural_2012, kao_coupled_2022} that make use of the principle of virtual work \cite{goldstein_classical_2002}, and integrating kinematic constraints to ensure that the solution is physically plausible.
Such methods have been shown to be equivalent to finite element methods (FEM) \cite{shin_reconciling_2016} for rigid objects, and, similar to FEM, are computationally expensive and non-deterministic.
Stability assessment under a large number of random force disturbances is perfomed in \cite{chen_planning_2021} by solving a \textit{min-max} optimization problem for every perturbation, making it the primary bottleneck of the planning algorithm.
We instead propose a physically sound heuristic that is quick to compute by avoiding combinatorial complexity.
Also, in contrast to \cite{chen_planning_2021}, our approach considers stability first instead of verifying it \rev{only} at the end.

Classical methods used to assess the stability of a grasp \cite{borst_fast_1999, ferrari_planning_1992} can provide a quantitative evaluation of the stability of a workpiece, while newer algorithms can produce stable multi-object grasps \cite{agboh_learning_2023} with rigid convex objects.
Although these methods offer a comparative basis for different placements, their application to object placement planning is limited. 
This is because they do not take into account the fact that fixed rigid objects can withstand very large forces, nor do they guide the planner on how to position an object on a stable structure---two aspects that our proposed algorithm addresses.

In general, the assembly task planning problem has been shown to be NP-complete \cite{kavraki_complexity_1993},
which explains why task and motion planning methods \cite{toussaint_logic-geometric_2015} have only been applied to small-scale problem instances.
Learning-based methods that do not resort to handcrafted heuristics have been proposed for assembly planning
\cite{xu_regression_2019, zhu_hierarchical_2020} but also struggle to generalize across tasks \cite{jiang_learning_2012}.
While most object planning algorithms ignore the inertial parameters of the objects, the work in \cite{haustein_object_2019} make use of the centre of mass of the object being placed but is limited to isolated placements on fixed horizontal surfaces.
\rev{Through our experiments, we show examples where this common approach may produce unstable placements.}
As an attempt at a more general object placement planning algorithm, physics simulators have been integrated into planning algorithms \cite{lee_object_2023} but are prohibitively slow to generate valid plans, even for small-scale problems.
In contrast, this work proposes a general method derived from first principles that avoids the overhead of dynamics simulators to better scale to larger problems.

\rev{Planning stable placements not only requires determining reaction forces but also the set of contact points forming the interfaces between objects, which is a complex task in general.}
\rev{Hence, previous works have simplified the problem by limiting the shape of the objects to rectangular workpieces \cite{chen_planning_2021, motoda_2022_shelf}, or by segmenting the scene into simple shapes like cylinders and cuboids \cite{kartmann_2018_extraction}.}
\rev{In contrast, our method can handle any shape described by a triangular mesh, with the placement planning time growing sub-quadratically with the number of vertices.}

\revv{In this work, we tackle the planning aspects of the inertia-aware object placement planning problem.}
\revv{However, a complete system would also include a perception pipeline to identify the properties of unknown objects.}
\revv{Particularly relevant to placement planning in static conditions, the shape and inertial parameters of a manipulated object are accurately identified in \cite{nadeau_sum_2023} with a commonly available force-torque sensor.}
\revv{Moreover, the approach in \cite{sundaralingam_2021_hand} uses tactile sensors to estimate the inertial parameters and the friction of manipulated objects, which could enable our method to deal with unknown objects.}

\section{Inertia-aware Object Placement Planning} %
\label{sec:inertia_aware_object_placement_planning}
\rev{Let $\mathcal{O}$ be a set of $N$ objects with known shape $\mathcal{V}_i\subset \Real^3$ and material composition $\rho_i(\Vector{p})~\forall~\Vector{p} \in \mathcal{V}_i$ for $i \in \left\{1,\dots,N\right\}$.}
\rev{Let an assembly $\mathcal{O}_a \subset \mathcal{O}$ be a set of objects in frictional contact, and $\mathcal{O}_f \subset \mathcal{O}_a$ be the subset of objects fixed in the limited space of the scene.}
\rev{The problem is to find a sequence of penetration-free placement poses $\Pose \in \LieGroupSE{3}$ that maximizes the force needed to displace any object.}
Hence, different from typical assembly planning problems, the goal configuration is not given in terms of the pose of the objects in the assembly, but rather in terms of an objective that is to be maximized. 

Unless stated otherwise, the RIGID notation convention \cite{nadeau_rigid_2024} is used and $\Pos{a}{b}{c}$ is the position vector of $\CSys{a}$ with respect to $\CSys{b}$ and expressed in $\CSys{c}$. The orientation of $\CSys{a}$ relative to $\CSys{b}$ is given by $\Rot{a}{b}$, and a unit-length vector $\VVec$ is denoted by $\hat{\VVec}$. The skew-symmetric operator $\Skew{\cdot}$ is defined such that $\Skew{\UVec}\VVec = \UVec\CrossP\VVec$ \revv{for the vectors $\UVec$ and $\VVec$ in $\mathbb{R}^3$}. \revv{An identity matrix of size $n\times n$ is denoted $\IdentityMatrix{n}$.}

\section{Static Robustness Assessment}
\label{sec:static_robustness_assessment}
When placing an object amongst others, it is usually desirable that the object be unlikely to move after being placed.
For rigid polyhedral objects, ultimately, instability will occur when an object slips or topples.
In both cases, the event is triggered when an applied force exceeds a threshold that we call the \textit{static robustness}: the maximum amount of force that can be exerted along a given direction, on a given point, before an object in the assembly moves.
Formally, the static robustness $r$ is defined as
\begin{equation}
    r: \mathbb{R}^3 \times \mathbb{S}^2 \rightarrow \mathbb{R}_+
\end{equation}
that maps a position $\Pos{}{}{} \in \mathbb{R}^3$ and direction $\EForce[\hat] \in \mathbb{S}^2$ to a scalar value $r$.
It follows that planning for stable placement involves reasoning about contact forces in the assembly.
However, computing frictional contact forces in an assembly has been shown to be NP-hard \cite{baraff_coping_1991}, and is made more complex by the existence of a coupling between forces and kinematic constraints \cite{kaufman_coupled_2009}.
\rev{In this work, we make the following assumptions about object dynamics:
\begin{enumerate}
    \item the Coulomb friction model is in effect,
    \item objects are very hard and only deform at contact points,
    \item contact points deform under the law of linear elasticity,
\end{enumerate}
with the two last assumptions being common with the Hertzian contact model \cite{handbook_contact_modeling}.}

\subsection{Solving for Reaction Forces}\label{sec:solving_reaction_forces}
In order to be quick to compute, a successful heuristic will need to avoid any computation whose complexity grows exponentially with the number of objects in the assembly.
Hence, we solve a linearly-constrained quadratic program \cite{whiting_structural_2012} that uses a pyramidal approximation of the friction cone, enforce static equilibrium, and avoid tensile forces.
This convex optimization problem can be solved in polynomial time \cite{baraff_analytical_1989} by standard solvers \cite{wachter_implementation_2006}, but may find unphysical solutions due to the approximation of the friction cone.
Focusing on a single object with $I$ contact points, let
\begin{align}
    \Matrix{A} &= \bbm \Matrix{A}_1 & \cdots & \Matrix{A}_I \ebm
    \text{ with }
    \Matrix{A}_i =
    \bbm
        \IdentityMatrix{3}\\
        \Skew{\Pos{i}{w}{w}}
    \ebm
    \rev{\CFrame{i}}
\end{align}
be a data matrix where $\Pos{i}{w}{w}$ is the position of the $i$-th contact point on the object and \rev{$\CFrame{i} = \bbm \UnitVec{u}_i & \UnitVec{v}_i & \UnitVec{n}_i \ebm$ defines a local frame with $\UnitVec{n}_i$ being the inward normal to the contact surface as pictured in \cref{fig:friction_cone}.}
\rev{Let}
\begin{equation}
    \label{eqn:gravity_wrench}
    \Vector{b} = -\Mass\bbm \IdentityMatrix{3} \\ \Skew{\Pos{c}{w}{w}} \ebm \Gravity
\end{equation}
be the wrench exerted by gravity on the centre of mass $\Pos{c}{w}{w}$ of the object whose mass is $\Mass$.
\rev{The} system of equations involving the forces $\rev{\Force = \bbm \RForce{1}{w}{w}\Tran, \cdots, \RForce{I}{w}{w}\Tran \ebm\Transpose}$ on the object is
\begin{equation}
    \Matrix{A} \Force = \Vector{b},
\end{equation}
where $\RForce{i}{w}{w}$ is the reaction force exerted at the $i$-th contact point and measured in the world frame.
The reaction force at the \rev{$i$-th contact point expressed in $\CFrame{i}$ is $\RForce{i}{w}{i}$} such that components $\RForce{i_n}{w}{i}$ and $\RForce{i_t}{w}{i}$ are respectively perpendicular and parallel to the contact surface.
\rev{When dealing with multiple objects, $\Matrix{A}$ and $\Vector{b}$ are constructed for every object and concatenated vertically.}
\rev{Note that, as per D'Alembert's principle, a dynamical system can be reduced to an equivalent static system by considering inertial forces that are due to the motion of the system.}
\rev{Including all inertial forces in \cref{eqn:gravity_wrench} would enable our method to handle dynamic scenarios as well.}

The principle of virtual work \cite{goldstein_classical_2002} dictates that the true distribution of forces should minimize the sum of squared forces for rigid objects in static equilibrium, such that
\begin{align}
    \label{eqn:optim_problem}
    &\min_{\Force \in\Real^{3n}} \quad \Force\Transpose\Force/2 \\
    &\quad\text{\emph{s.t.}} \quad \Matrix{A} \Force = \Vector{b}, \notag\\
    &\quad\quad\quad \Norm{\RForce{i_n}{w}{i}}_2 \geq 0, \hspace{1.8cm}\forall i \in \left\{1,\dots,I\right\}\notag\\
    &\quad\quad\quad \Norm{\RForce{i_t}{w}{i}}_1 \leq \mu_i \Norm{\RForce{i_n}{w}{i}}_2, \quad\forall i \in \left\{1,\dots,I\right\}\notag
\end{align}
\rev{determines} reaction forces with $\mu_i$ being the friction coefficient at the \rev{$i$}-th contact point.
The optimization problem in \cref{eqn:optim_problem} is a linearly constrained quadratic program (QP), for which the optimal solution can be obtained with standard solvers \cite{wachter_implementation_2006} if it exists.
The constraints in \cref{eqn:optim_problem} respectively enforce the equilibrium of the assembly, the compressive nature of the contact forces, and a pyramidal approximation of the Coulomb friction cone.

Alternatively, the minimal two-norm solution can be obtained via a QR decomposition \cite{golub_matrix_2013} of the data matrix, with
\begin{equation}
    \Matrix{Q}, \Matrix{R} = \text{qr}\left(\Matrix{A}\Transpose\right),
\end{equation}
by solving 
\begin{equation}
    \Matrix{R}_{1:6, 1:6}\Transpose \Vector{z} = \Vector{b}
\end{equation}
for $\Vector{z}$, where $\Matrix{R}_{1:6, 1:6}$ is the upper-left $6\times 6$ sub-matrix of $\Matrix{R}$.
The reaction forces can then be obtained with
\begin{equation}
    \label{eqn:reaction_forces_solution}
    \Force = \Matrix{Q}_{:, 1:6} \Vector{z},
\end{equation}
where $\Matrix{Q}_{:, 1:6}$ is built from the first six columns of $\Matrix{Q}$.
This approach is expected to be the fastest but is unconstrained and might violate the friction cone and the non-tensile force constraints.
For placement planning, the QR-based approach can be used during the iterative process by checking the magnitudes of the constraint violations and verifying the validity of promising solutions with the optimization-based approach.
\rev{However, due to static indeterminacies, the existence of a solution to the optimization problem in \cref{eqn:optim_problem} does not guarantee that the assembly is stable \cite{pang_stability_2000}. Stability can only be guaranteed by testing all possible solutions.}
\rev{If needed, a different method could be used to compute reaction forces, while still seamlessly integrating with our proposed planner.}

\subsection{Robustness to Slipping}

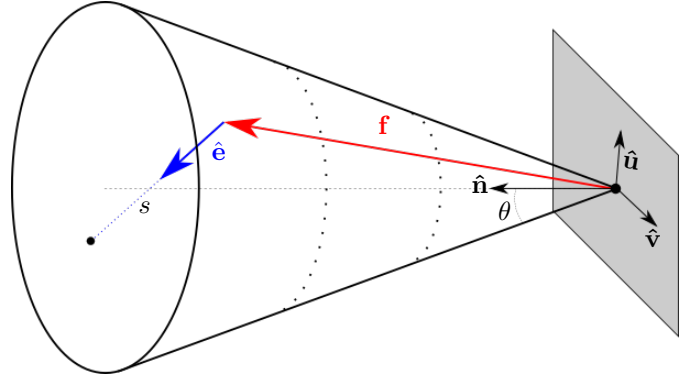
\begin{figure}[h]
    \centering
    \begin{overpic}[width=0.9\linewidth]{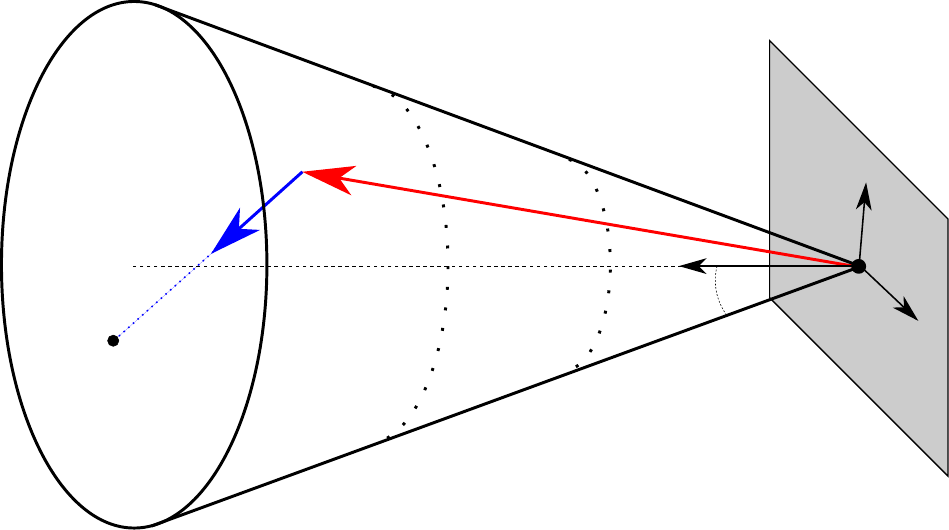}
        \put(55,36){$\color{red}\RForce$}
        \put(30,32){$\color{blue}\EForce[\hat]$}
        \put(73,23){$\theta$}
        \put(19,24){$s$}
        \put(91.5,31){\rotatebox{-10}{$\UnitVec{u}$}}
        \put(95,19){$\UnitVec{v}$}
        \put(69,27){$\UnitVec{n}$}
    \end{overpic}
    \caption{\rev{Friction cone at a contact point with reaction force $\RForce$ (red) and an external force $\EForce[\hat]$ (blue). Increasing the magnitude of the external force vector to $\Mag$ will produce an intersection with the boundary of the cone at $\RForce+\Mag\thinspace\EForce[\hat]$.} 
}
    \label{fig:friction_cone}
\end{figure}

\rev{Consider the friction cone at a given contact point, defined by its local frame $\CFrame{i} = \bbm \UnitVec{u}_i & \UnitVec{v}_i & \UnitVec{n}_i \ebm$ and angle $\theta$, as shown in \cref{fig:friction_cone}.}
\rev{Let $\RForce$ and $\EForce[\hat]$ be a reaction force vector and a unit external force vector at the contact point, respectively.}
The maximal magnitude of the external force vector $\EForce=\Mag\thinspace\EForce[\hat]$ that can be applied to the contact point is the one that produces an intersection between the friction cone and the line defined by the external force vector.

\rev{The general equation defining a circular cone embedded in 3D space and supported by angle $\theta$ is given by}
\begin{align}
    \label{eq:cone_equation}
    \tan^2(\theta) z^2 = x^2 + y^2,
\end{align}
where $\tan^2\theta = \mu^2$, the square of the friction coefficient at the contact point.
\rev{Let $\EForce[\hat]{i}{w}{i} = \bbm e_u & e_v & e_n\ebm\Transpose$ and $\RForce{i}{w}{i} = \bbm r_u & r_v & r_n \ebm\Transpose$ be the components of the forces expressed in $\CFrame{i}$, and $\Mag$ be the magnitude of the external force vector.}
\rev{A point on the cone boundary can be given as a function of the forces at the contact point with}
\begin{align}
    \label{eq:line_equation}
    x &= se_u + r_u,\\
    y &= se_v + r_v,\\
    z &= se_n + r_n,
\end{align}
as shown in \cref{fig:friction_cone}.
Substituting \cref{eq:line_equation} into \cref{eq:cone_equation} yields
\begin{align}
    \label{eq:cone_line_equation}
    &\rev{\tan^2(\theta) (se_n + r_n)^2 = (se_u + r_u)^2 + (se_v + r_v)^2}%
\end{align}
which is a quadratic in $\Mag$,
\begin{align}
    \label{eq:quadratic_equation}
    0 = as^2 + bs + c
\end{align}
where
\begin{align}
    \label{eq:quadratic_coefficients}
    a &= \mu^2e_n^2-e_u^2-e_v^2,\\
    b &= 2\left(\mu^2r_ne_n-r_ue_u-r_ve_v\right),\\
    c &= \mu^2r_n^2-r_u^2-r_v^2,
\end{align}
and for which a particular solution is given by
\begin{align}
    \label{eq:quadratic_solutions}
    s = \frac{-b-\sqrt{b^2-4ac}}{2a},
\end{align}
the well-known quadratic formula.

Hence, the static robustness $r$ of a contact point is given by 
\begin{align}
    r &= \begin{cases}
        s & \text{if } s > 0\\
        \infty & \text{otherwise}.
    \end{cases}
\end{align}
The static robustness is infinite if $\EForce[\hat] + \RForce$ does not intersects with the friction cone.
Since the reaction forces are not constrained to obey the non-linear Coulomb friction model in the computation of \cref{eqn:reaction_forces_solution}, it is possible that 
\begin{align}
    \mu\Norm{\RForce{i_n}{w}{i}} &< \Norm{\RForce{i_t}{w}{i}}\\
    \mu^2r_n^2 &< \left(r_u^2+r_v^2\right)\\
    c &< 0 ,
\end{align}
which implies that the contact point is not holding. 
In such a case, or similarly if the radicand $b^2-4ac$ is negative, the static robustness of the contact point is set to zero.
\rev{Assuming that contact points obey the laws of linear elasticity, the stress incurred at each contact point is cumulated to produce the total frictional force \cite{sinha_contact_1990, bhushan_contact_1998}.}
\rev{Hence, the robustnesses of several contact points are summed with}
\begin{equation}
    \label{eqn:slipping_robustness_sum}
    \revv{r_\text{slip} = \sum r_i \hspace{0.3cm} \forall i \in \left\{1, \cdots, I\right\}}
\end{equation}
to produce the robustness of the object to slipping, where $r_i$ is the robustness of the $i$-th contact point.

\subsection{Robustness to Toppling}
\begin{figure}[t]
    \centering
    \begin{overpic}[width=0.8\linewidth]{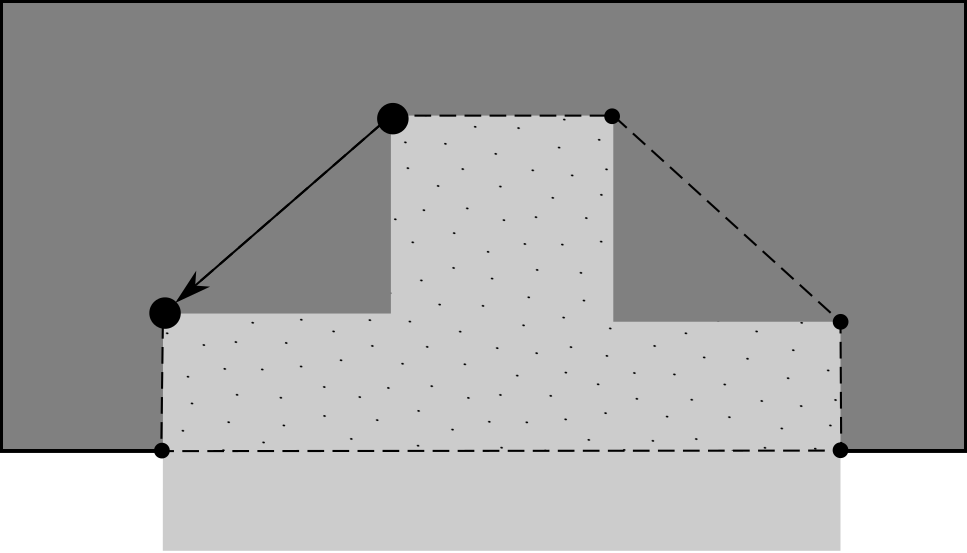}
        \put(25.5,35){$\color{white}\Axis$}
        \put(10.5,25){$\color{white}k_t$}
        \put(35.5,47){$\color{white}k_s$}
    \end{overpic}
    \caption{\rev{View from above of a non-convex object (light grey) atop another object (dark grey). The edges of the convex hull (dashed line) of the contact points are potential toppling axes.}}
    \label{fig:toppling_axis}
\end{figure}

The robustness of an object to being pushed over is given by the minimum amount of force required to rotate the object about some axis $\Axis$, defined by the line joining a pair of contact points \rev{as shown in \cref{fig:toppling_axis}.}
For an isolated object being pushed, the toppling axis is guaranteed to be an edge of the convex hull of the contact points.
\rev{Any} other axis would require a greater amount of force to rotate the object because the lever arm would be smaller.
Hence, \rev{the convex hull of the $n$ contact points} is computed via the QuickHull algorithm \cite{barber_quickhull_1996}, whose complexity is $O(n^2)$ in the worst case, to produce a list of axes about which the object could topple.

\rev{Let, $\Pos{c}{k_s}{w}$ be the location of the centre of mass relative to $k_s$ and expressed in the world frame, and $\Mass\Gravity$ be the weight of the object.}
\rev{An axis is a \emph{valid} (i.e., possible) toppling axis if each contact point normal exerts a positive torque about the axis, which is verified with}
\begin{equation}
    \left(\Pos{i}{k_s}{w}\CrossP\RForce{i_n}{w}{w}\right)\cdot\Axis > 0 \hspace{2.5mm}\forall i \in \left\{1, \cdots, I\right\},
\end{equation}
\rev{and that the weight of the object exerts a negative torque about the axis, which is verified with}
\begin{equation}
    \left(\Pos{c}{k_s}{w}\CrossP\Mass\Gravity\right)\cdot\Axis < 0.
\end{equation}
The robustness of an object to toppling about $\Axis$ is given by 
\begin{equation}
    s_i = \frac{\left(\Pos{c}{k_s}{w}\CrossP\Mass\Gravity\right)\cdot\Axis}{\left(\Pos{i}{k_s}{w}\CrossP\EForce[\hat]\right)\cdot\Axis},
\end{equation}
which is the force required to compensate for the torque due to gravity that is holding the object in place.  
The maximum amount of force that can be exerted along a given direction on a given point before the object topples is given by
\begin{equation}
    r_\text{top} = \min\left\{s_i\right\} \quad \forall i \in \left\{1, \cdots, I\right\},
\end{equation}
\rev{which provides the minimum toppling robustness} when all valid axes are considered.

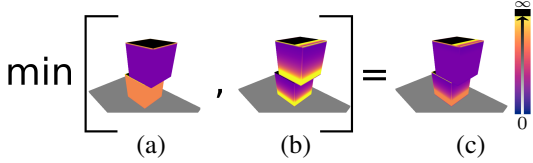
\begin{figure}[t]
    \centering
    \begin{overpic}[width=0.8\linewidth]{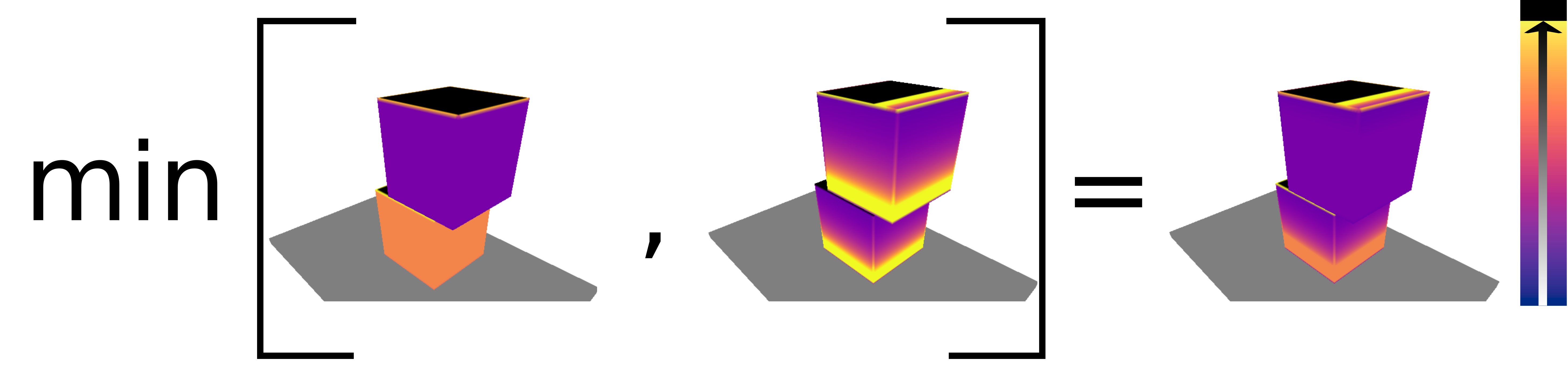}
        \put(26,-3){(a)}
        \put(53,-3){(b)}
        \put(86,-3){(c)}
        \put(97.4,1.5){$\scriptstyle{0}$}
        \put(96.8,24.5){$\scriptscriptstyle{\infty}$}
    \end{overpic}
    \vspace{0.2cm}
    \caption{Combining slipping (a) and toppling (b) robustnesses to obtain the overall static robustness (c) \rev{with values colour-coded according to the scale on the right (black being infinite).}}
    \label{fig:combining_robustnesses}
\end{figure}

\rev{Slipping and toppling robustnesses may be combined as}
\begin{equation}
    \label{eqn:overall_robustness}
    r = \min\left\{r_\text{slip}, r_\text{top}\right\}
\end{equation}
\rev{to obtain the overall static robustness of the object, as pictured in \cref{fig:combining_robustnesses}.}
With \cref{eqn:overall_robustness}, the definition of the static robustness $r$ as a mapping from a direction $\EForce[\hat]$ and position $\Pos{p}{w}{w}$ to a scalar value $r$ is fulfilled.
The result of the mapping can be seen in \cref{fig:frontpage}, \cref{fig:combining_robustnesses}, and \cref{fig:scenes}, where the colour indicates the relative magnitude of the maximal force that can be applied on a point \rev{in the direction normal to the surface.}
A lighter colour indicates that a greater force can be applied before the object slips or topples, with black indicating that an infinite force can be applied.
\rev{The} robustness function obtained via the use of our heuristic is unlikely to be exact in multi-objects assemblies.
\rev{However,} it is quick to compute because it avoids any operation whose computational complexity grows exponentially with the number of objects in the assembly.

\section{Placement Planning from Static Robustness}
\label{sec:placement_planning_from_static_robustness}

Ideally, the normal contact force at each contact point is much greater than the tangential force such that the object is far from slipping.
Therefore, our proposed planner uses the robustness to normal forces at surface points to plan stable placements.
In the following, we assume that a \rev{known} object has to be placed in a given assembly of objects for which the robustness in the normal direction at all points on the surface of the objects in the assembly is known.
\rev{Also, placements are assumed to be performed in a linear (i.e., a single object is moved at a time), monotone (i.e., placing does not involve rearranging), and sequential (i.e., a single gripper is used) fashion \cite{wang_state_2021}.}

By placing an object in an assembly, at least one contact interface between objects is created\rev{, which fixes some of the degrees of freedom (DoFs) of the object.}
Any nondegenerate\footnote{In \cite{xiao_automatic_1993}, edge/vertex, vertex/vertex, and parallel edge/edge contacts are defined as being degenerate.} contact interface can be categorized into \cite{xiao_automatic_1993, ji_planning_2001}:
\begin{itemize}
    \item a face/face plane contact that fixes 3 DoFs,
    \item a face/edge line contact that fixes 2 DoFs,
    \item an edge/edge point contact that fixes 1 DoF, and
    \item a face/vertex point contact that fixes 1 DoF.
\end{itemize}
At least three contact interfaces are necessary to constrain the object to a specific pose by fixing all 6 DoFs \cite{ji_planning_2001}.
However, to place an object successfully without inducing a collision, at least one DoF must remain unconstrained.
With two non-collinear contact interfaces, from one (e.g., with face/face interfaces) to four (e.g., with face/vertex interfaces) DoFs will remain unconstrained, \rev{ensuring that the object can be placed without inducing a collision.}
A third contact interface might be needed \rev{to ensure} stability under gravity, but \rev{the interface} does not necessarily need to be selected prior to the computation of the placement pose.
Furthermore, in a face/face contact situation, \rev{the normal vectors on each side of the contact interface must be opposed.} 
\rev{In a face/edge contact, the normal vector of the face must be orthogonal to the contact edge.}

Our approach is summarized by \cref{algo:planning}, with each step described in greater detail in the following sections.
\rev{Our algorithm is based on sampling points on the assembly and on the object to place, and then defining a pose for the object based on the two sets of points.}
With this sampling-based approach, it is expected that many candidates will turn out to be invalid, and that the process will need to be iterated several times before a stable, non-penetrating pose is found.
Each iteration therefore needs to be performed as quickly as possible.
\begin{algorithm}[t]
    \DontPrintSemicolon
    \LinesNumbered
    \SetKwInOut{Input}{input}\SetKwInOut{Output}{output}
    \Input{\rev{Assembly $\mathcal{O}_a$, object to place $\mathcal{O}_i$,\\ static robustness map $\mathrm{r}(\Vector{p}_i, \Normal[\hat]{i})~~\forall~\Vector{p}_i\in\mathcal{P}$}}
    \Output{Stable pose $\Pose{o}{w}{}$ of $\mathcal{O}_i$ if found}
    \While{max. number of iterations not reached}{
        Sample two points $\Pos{a}{w}{w}$ and $\Pos{b}{w}{w}$ on $\mathcal{O}_a$ with \cref{eqn:prob_dist_with_robustness}\;
        Find two points $\Pos{q}{o}{o}$ and $\Pos{r}{o}{o}$ on the support geometry of $\mathcal{O}_i$ whose normals oppose those on the assembly and whose relative positions coincide\;
        Define $\Pose{o}{w}{}$ from $\Pos{a}{w}{w}$, $\Pos{b}{w}{w}$, $\Pos{q}{o}{o}$, and $\Pos{r}{o}{o}$\;
        \If{$\Pose{o}{w}{}$ does not result in inter-penetration}{
            Solve \cref{eqn:reaction_forces_solution} for reaction forces (QR-based)\;
            \If{tension forces are below threshold}{
                Solve \cref{eqn:optim_problem} for reaction forces (QP-based)\;
                \If{forces are equilibrated}{
                    \Return $\Pose{o}{w}{}$\;
                }
            }
        }
    }
    \caption{Placement Planning From SR}
    \label{algo:planning}
\end{algorithm}

\subsection{Contact Point Selection}
The selection of candidate contact points on the surface of objects in the assembly is performed in an iterative, probabilistic manner based on the robustness of the objects to normal forces.
At each iteration of the planning algorithm, two points are sampled according to the probability distribution in \cref{eqn:prob_dist_with_robustness}.
\rev{With $\mathrm{r}(\Vector{p}, \Normal[\hat])$ being the robustness of the object in the normal direction at point $\Vector{p}$, the probability of sampling this point is given by}
\begin{equation}
    \label{eqn:prob_dist_with_robustness}
    \mathrm{P}\left(\Vector{p}\right) = \frac{\min\left(\mathrm{r}(\Vector{p}, \Normal[\hat]), Q\right)}{\sum\limits_{\Vector{p}_i\in\mathcal{P}}\min\left(\mathrm{r}(\Vector{p}_i, \Normal[\hat]{i}), Q\right)},
\end{equation}
where $Q$ is the robustness above which the odds of being sampled are set to be constant.
In practice, $Q$ can be defined such that the largest finite robustness in the assembly has a small probability of being sampled (e.g., $10\%$).
\rev{This can be done} by initializing $Q$ to a larger value $Q_0 > 0$ that is exponentially decayed with the number of iterations $k$ at a rate $\lambda < 1$ with $Q_k = Q_0 \lambda^k$.
\rev{This way,} the probability of sampling any point in $\mathcal{P}$ converges to
\begin{equation}
    \lim_{k\to\infty} \mathrm{P}\left(\Vector{p}\right) = \frac{Q_k}{\vert\mathcal{P}\vert Q_k} = \frac{1}{\vert\mathcal{P}\vert},
\end{equation}
where $\vert\mathcal{P}\vert$ is the cardinality of $\mathcal{P}$, the set of surface contact points.
With such a scheme, the odds of sampling a point with a large robustness are initially higher, but decrease with the number of iterations.
\rev{All} points are eventually considered with almost equal probability, and any valid placement pose is eventually considered.

\subsubsection{Extension: Sampling on Fixed Supports}
Sometimes, it might be desirable to place the object on a fixed support (e.g. table) instead of amongst the objects in the assembly.
For instance, if placing the object on the assembly would result in a weakened assembly, it might be preferable to place the object on a fixed support whose static robustness is, by definition, infinite.
In this case, however, it is usually desired that the object be placed in a compact manner.
The probability function in \cref{eqn:prob_dist_with_robustness} can be easily extended to accommodate this requirement. %
\rev{For instance, the set of scene points $\mathcal{P}$ can be augmented with a set of points $\mathcal{P}_f$ uniformly distributed on the fixed support with}
\begin{align}
    w\left(\Vector{p}\right) &=
    \begin{cases}
        P(\Vector{p})~\forall \Vector{p}\in\mathcal{P}\\
        \text{max}_{\Vector{p}\in\mathcal{P}}\left(\mathrm{P}\left(\Vector{p}\right)\right)~\forall \Vector{p}\in\mathcal{P}_f\\
    \end{cases},\\
    \label{eqn:unnorm_prob_dist_with_robustness_fixed}
    w\left(\Vector{p}\right) &= w\left(\Vector{p}\right) e^{-\frac{\gamma^k}{s^2}\Norm{\Pos{p}{w}{w} - \Pos{c}{w}{w}}^2}~\forall \Vector{p}\in\mathcal{P}\cup\mathcal{P}_f,\\
    \label{eqn:prob_dist_with_robustness_fixed}
    P\left(\Vector{p}\right) &= \frac{w\left(\Vector{p}\right)}{\sum w\left(\Vector{p}\right)}~\forall \Vector{p}\in\mathcal{P}\cup\mathcal{P}_f,
\end{align}
\rev{where \cref{eqn:unnorm_prob_dist_with_robustness_fixed} increases the probability of sampling points closer to the centroid $\Pos{c}{w}{w}$ of the scene and \cref{eqn:prob_dist_with_robustness_fixed} ensures that the sum of the probabilities is equal to one.}
In \cref{eqn:unnorm_prob_dist_with_robustness_fixed}, $\gamma$ is the rate at which the probability of sampling a point close to the centroid of the scene decreases and $s$ is the scale of the scene.

\subsection{Object Point Selection}
To define a placement pose, a pair of object features (i.e., faces or edges) is considered.
One point per feature is selected such that the distance between the object points corresponds to the \rev{distance} between the scene contact points.
To favour stability, the object points are selected such that the line joining the points is as close as possible to the centre of mass of the object.
Prior to point selection, candidate pairs of features on the object are validated by considering the relative orientation of the feature normals to ensure that they can \textit{afford} the support geometry of the contact points in the assembly.
\rev{Each feature pair can either be \textit{face-face} (parallel, coplanar, or intersecting), \textit{face-edge} (parallel, or intersecting), or \textit{edge-edge} (parallel, intersecting, or skew).}

In the following, we assume that the contact points $\Pos{a}{w}{w}$ and $\Pos{b}{w}{w}$ are sampled on object faces in the scene with normals $\Normal[\hat]{a}{w}$ and $\Normal[\hat]{b}{w}$ such that $\Norm{\Pos{a}{w}{w}-\Pos{b}{w}{w}} = L$.
The relative orientation of $\Normal[\hat]{a}{w}$ and $\Normal[\hat]{b}{w}$ can be obtained with
\begin{align}
    \AAxis &= \Normal[\hat]{a}{w} \CrossP \Normal[\hat]{b}{w} ,\\
    \AAngle & = \arccos\left(\Normal[\hat]{a}{w} \DotP \Normal[\hat]{b}{w}\right) ,\\
    \Rot{b}{a} &= \IdentityMatrix{3} + \sin\left(\AAngle\right)\Skew{\AAxis} + \left(1-\cos\left(\AAngle\right)\right)\Skew{\AAxis}^2,
\end{align}
where $\AAxis$ is the axis of rotation from $\Normal[\hat]{a}{w}$ to $\Normal[\hat]{b}{w}$, $\AAngle$ is the angle of rotation about $\AAxis$ from $\Normal[\hat]{a}{w}$ to $\Normal[\hat]{b}{w}$, and $\Rot{b}{a}$ is the rotation matrix from $\Normal[\hat]{a}{w}$ to $\Normal[\hat]{b}{w}$.

\begin{figure}[h]
    \begin{overpic}[width=1\linewidth]{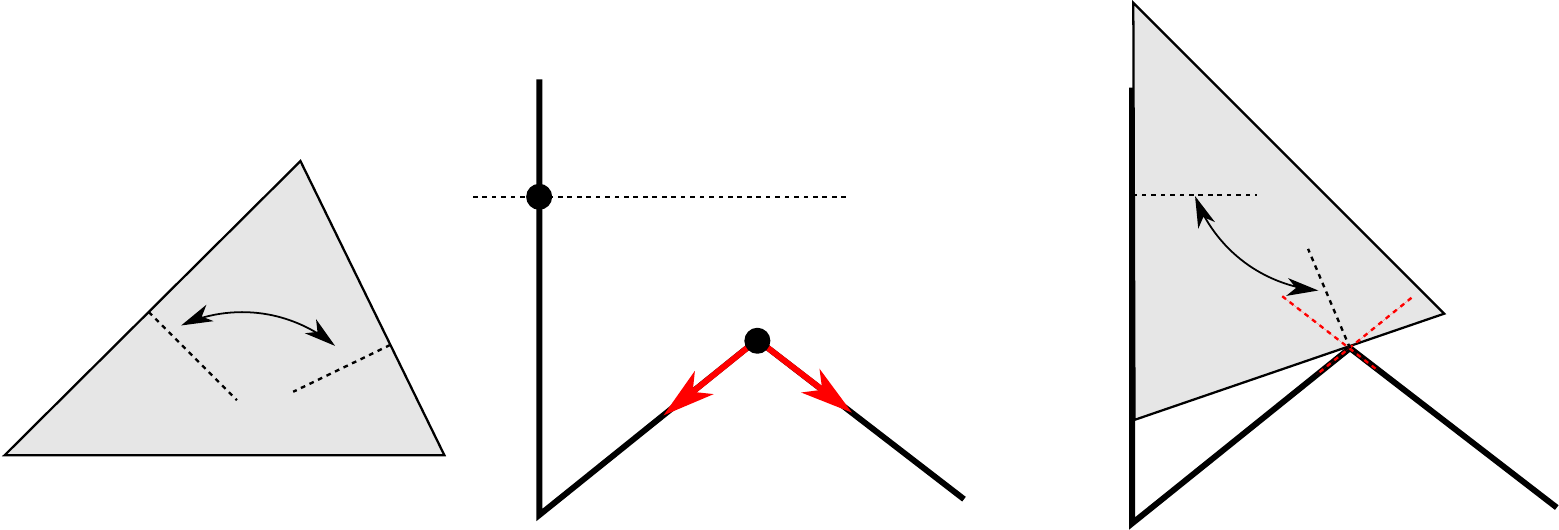}
        \put(6,-3){(a) Object}
        \put(40,-3){(b) Scene}
        \put(70,-3){(c) Object Afforded}

        \put(6,10){$\Normal[\hat]{f_1}{o}$}
        \put(20,7){$\Normal[\hat]{f_2}{o}$}
        \put(12,16){$\Rot{\UnitVec{n}_2}{\UnitVec{n}_1}$}

        \put(35,22){$b$}
        \put(49,13){$a$}

        \put(51,11.5){\color{red}$\Side[\hat]{e_2}{w}$}
        \put(38.5,11.5){\color{red}$\Side[\hat]{e_1}{w}$}

        \put(40,223){$\Normal[\hat]{b}{w}$}

        \put(72.5,15){$\Rot{\UnitVec{n}_2}{\UnitVec{n}_1}$}
        
    \end{overpic}
    \vspace{0mm}
    \caption{\rev{Affordance test to see if $\Normal[\hat]{f_2}{o}$ would lie between $\Side[\hat]{e_1}{w}$ and $\Side[\hat]{e_2}{w}$ when $\Normal[\hat]{f_1}{o}$ opposes $\Normal[\hat]{b}{w}$.}}
    \label{fig:scene_affording_object}
\end{figure}

\textbf{Validating Feature Pairs}
A face will afford another face if its normal is parallel and opposed to the normal of the other face.
In 3D, an edge $\Edge{i}{o}$ joins two adjacent faces whose normals are denoted by $\Normal[\hat]{e_1}{o}$ and $\Normal[\hat]{e_2}{o}$.
The vectors directed from the edge to the inside of the face, along the face and orthogonal to the edge are given by %
\begin{align}
    \Side[\hat]{e_1}{o} &= \left( \Edge{i}{o} \CrossP \Normal[\hat]{e_1}{o} \right) / \Norm{\Edge{i}{o} \CrossP \Normal[\hat]{e_1}{o}},\\
    \Side[\hat]{e_2}{o} &= \left( \Edge{i}{o} \CrossP \Normal[\hat]{e_2}{o} \right) / \Norm{\Edge{i}{o} \CrossP \Normal[\hat]{e_2}{o}},
\end{align}
\rev{and termed the \textit{side vectors} (see \cref{fig:scene_affording_object}b).}
\revv{The object normals and side vectors are computed in the object frame, and the pose of the objects in the scene is used to project these vectors in the world frame, accommodating dynamically changing scenes.}

A face with outward normal $\Normal[\hat]{}{o}$ will afford an edge if the normal is orthogonal to the edge direction $\UVec{e}{o}$ such that
\begin{equation}
    \UVec{e}{o} \DotP \Normal[\hat]{}{o} = 0 ,
\end{equation}
and if the normal lies between the two side vectors of the edge.
Intuitively, a planar surface with infinitesimal area should touch the edge without touching the faces on either side.
A simple test to check if the normal lies between $\Side[\hat]{e_1}{o}$ and $\Side[\hat]{e_2}{o}$ is defined with
\begin{align}
    u &= \bbm 0&0&1\ebm \left( \Side[\hat]{e_1}{o} \CrossP \Side[\hat]{e_2}{o} \right),\\
    v &= \bbm 0&0&1\ebm \left( \Normal[\hat]{}{o} \CrossP \Side[\hat]{e_2}{o} \right),\\
    w &= \bbm 0&0&1\ebm \left( \Normal[\hat]{}{o} \CrossP \Side[\hat]{e_1}{o} \right),
\end{align}
such that the normal lies between the two side vectors if $uv \leq 0$ and $uw \geq 0$.

Since the normals sampled in the scene are expressed in a different frame than the ones on the object, the components of the normal vectors cannot be simply compared.
\rev{Instead, the relative orientation of the object normals is considered to check if the scene can afford the object normals.}
\rev{For instance, assuming that the normal to the first feature selected on the object opposes $\Normal[\hat]{b}{w}$, the normal of the second feature needs to lie between the side vectors around point $\Pos{a}{w}{w}$ when $\Normal[\hat]{f_1}{o}$ opposes $\Normal[\hat]{b}{w}$, as shown in \cref{fig:scene_affording_object}.}

\revv{\textbf{Selecting Object Points}}
Several cases are considered depending on the relative orientation of the features.
In each case, the points on the object are selected such that the distance between them is $L$ and that the line joining them is as close as possible to the centre of mass of the object.

\begin{figure}
    \centering
    \begin{overpic}[width=0.8\linewidth]{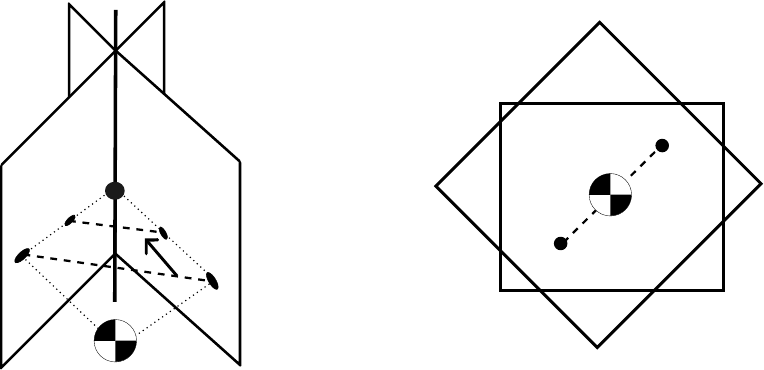}
        \put(13,-5){(a)}
        \put(-2.5,30){$P_1$}
        \put(28,30){$P_2$}
        \put(11.8,48.5){$\Edge[\hat]{i}{o}$}
        \put(0.5,11){$o_1$}
        \put(27,8){$o_2$}
        \put(15.3,24.8){$o_0$}
        \put(7.5,21){$q$}
        \put(21.5,18){$r$}
        \put(12,16.5){\tiny{$L$}}
        \put(76,-5){(b)}
        \put(75.5,6){$P_1$}
        \put(89.5,11.5){$P_2$}
        \put(72,18.5){$q$}
        \put(86,30.5){$r$}
        \put(76,15){$\frac{L}{2}$}
        \put(83,22){$\frac{L}{2}$}
    \end{overpic}
    \vspace{0.25cm}
    \caption{Face-face pair with (a) intersecting faces and (b) coplanar faces.}
    \label{fig:face_face_pair}
\end{figure}
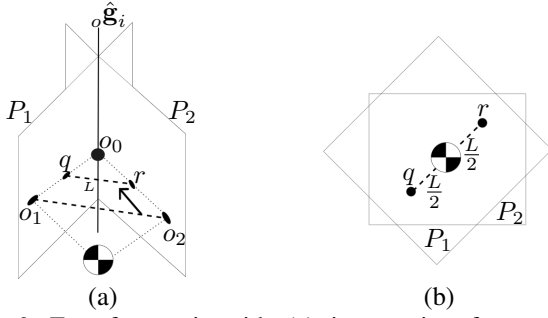

\textbf{A. Face-Face Pair}
Two faces sampled on the object can be parallel, coplanar, or intersecting, as shown in \cref{fig:face_face_pair}.

Let two planes be defined by
\begin{align}
    P_1(d_1) &: \Normal[\hat]{f_1}{o} \DotP \Vector{p} = d_1 ,\\
    P_2(d_2) &: \Normal[\hat]{f_2}{o} \DotP \Vector{p} = d_2 ,
\end{align}
where $\Vector{p}$ is a point on the plane, $\Normal[\hat]{f}{o}$ is the normal of the face plane, and $d$ is the distance of the plane from the origin.
If the faces are parallel but not coplanar, the distance between the faces $\left(d_1-d_2\right)\Normal[\hat]{f_2}{o}$ would need to be exactly equal to $L$ for the match to be valid.
\rev{This} is almost never the case and would produce infinitesimal interpenetrations (i.e., friction) when placing the object.

\textbf{A.1. Coplanar Faces}
If the faces are coplanar, there is an infinite number of point pairs that are distanced by $L$.
Amongst the possible pairs, we select one that minimizes the distance between the centre of mass of the object and the line joining the two points to maximize the odds of sampling a stable pose.
Let $\UVec{}{o}$ be in the direction given by the extent of the object projected onto the common plane of the two faces.
\rev{Selecting} points along the direction vector that are at a distance $\pm L/2$ from the projection of the centre of mass onto the face plane is done with
\begin{align}
    \label{eqn:selection_coplanar_faces}
    \Pos{o_1}{o}{o} &= d_1\Normal[\hat]{f_1}{o} + \Pos{c}{o}{o}-\left(\Pos{c}{o}{o} \DotP \Normal[\hat]{f_1}{o}\right)\Normal[\hat]{f_1}{o} ,\\
    \Pos{q}{o}{o} &= \Pos{o_1}{o}{o} + \frac{L}{2}\UVec{}{o} ,\\
    \Pos{r}{o}{o} &= \Pos{o_1}{o}{o} - \frac{L}{2}\UVec{}{o} ,
\end{align}
where $\Pos{c}{o}{o}$ is the centre of mass of the object expressed in the object frame.
In \cref{eqn:selection_coplanar_faces}, $\Pos{o_1}{o}{o}$ is the projection of the centre of mass onto the face plane, while $\Pos{q}{o}{o}$ and $\Pos{r}{o}{o}$ are respectively the point on the first and second face being considered.

\textbf{A.2. Intersecting Faces}
If the faces are intersecting, they share a common edge whose direction is given by
\begin{align}
    \Edge[\hat]{e}{o} &= \Normal[\hat]{f_1}{o} \CrossP \Normal[\hat]{f_2}{o} / \Norm{\Normal[\hat]{f_1}{o} \CrossP \Normal[\hat]{f_2}{o}} ,
\end{align}
and whose origin $\Pos{e}{o}{o}$ is defined by
\begin{align}
    \UVec &= \Normal[\hat]{f_1}{o} \CrossP \Edge[\hat]{e}{o} ,\\
    \Pos{e}{o}{o} &= \Pos{o_1}{o}{o} + \frac{d_2-\Normal[\hat]{f_2}{o} \DotP \Pos{o_1}{o}{o}}{\Normal[\hat]{f_2}{o} \DotP \UVec}\UVec ,
\end{align}
where $\Pos{o_1}{o}{o}$ is the projection of the centre of mass onto the plane of the first face, as defined previously.
The plane orthogonal to the common edge and passing through the centre of mass of the object intersects $\Edge[\hat]{e}{o}$ at
\begin{equation}
    \Pos{o_0}{o}{o} = \Pos{e}{o}{o} + \left(\Pos{c}{o}{o} \DotP \Edge[\hat]{e}{o}\right)\Edge[\hat]{e}{o}
\end{equation}
such that a triangle is formed by $\Pos{o_0}{o}{o}$, $\Pos{o_1}{o}{o}$, and $\Pos{o_2}{o}{o}$ in the plane passing through $\Pos{c}{o}{o}$.
A similar triangle connecting $\Pos{o_0}{o}{o}$, $\Pos{q}{o}{o}$, $\Pos{r}{o}{o}$ whose leg joining both faces has length $L$ can be defined with
\begin{align}
    \label{eqn:similar_triangle_finding_points}
    \Pos{q}{o}{o} &= \Pos{o_0}{o}{o} + \frac{L}{\Norm{\Pos{o_1}{o}{o}-\Pos{o_2}{o}{o}}}\left(\Pos{o_1}{o}{o}-\Pos{o_0}{o}{o}\right) ,\\
    \Pos{r}{o}{o} &= \Pos{o_0}{o}{o} + \frac{L}{\Norm{\Pos{o_1}{o}{o}-\Pos{o_2}{o}{o}}}\left(\Pos{o_2}{o}{o}-\Pos{o_0}{o}{o}\right) ,
\end{align}
where $\Pos{o_2}{o}{o}$ is the projection of the centre of mass onto the plane of the second face, and $\Pos{q}{o}{o}$ and $\Pos{r}{o}{o}$ are respectively the point on the first and second face being considered.

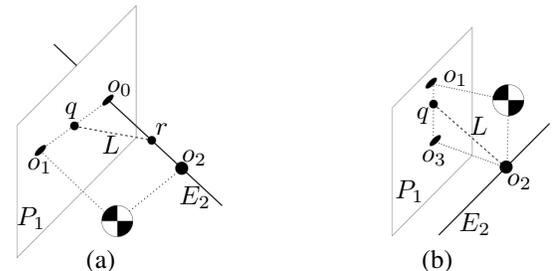
\begin{figure}[b]
    \centering
    \begin{overpic}[width=0.8\linewidth]{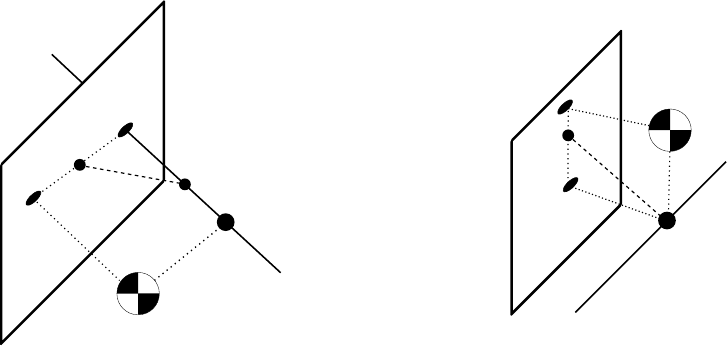}
        \put(13,-2){(a)}
        \put(16,19.5){$L$}
        \put(0,6){$P_1$}
        \put(30.5,10){$E_2$}
        \put(17,31){$o_0$}
        \put(2,16.5){$o_1$}
        \put(31,18.5){$o_2$}
        \put(9,27){$q$}
        \put(26,23){$r$}
        \put(76,-2){(b)}
        \put(85.5,23){$L$}
        \put(71,11){$P_1$}
        \put(83,5){$E_2$}
        \put(80,33){$o_1$}
        \put(92,13.5){$o_2$}
        \put(76,18){$o_3$}
        \put(75,26){$q$}
    \end{overpic}
    \vspace{0.1cm}
    \caption{Face-edge pair: (a) intersecting, (b) parallel.}
    \label{fig:face_edge_pair}
\end{figure}

\textbf{B. Face-Edge Pair}
In the situation where one interface is between two faces and the other interface is between a face and an edge, the line vector connecting the two contact points on the object will join a point on the face to a point on the edge, as shown in \cref{fig:face_edge_pair}.
Let the plane of the face/face interface be defined by
\begin{equation}
    \rev{P_1(d) : \Normal[\hat]{f}{o} \DotP \Pos{q}{o}{o}= d} ,
\end{equation}
and the edge be defined by
\begin{equation}
    \rev{E_2(t) : \Pos{r}{o}{o} = \Pos{e}{o}{o} + t \UVec{e}{o}} ,
\end{equation}
\rev{such that $\Vector{q}$ is constrained to lie on the plane of the face/face interface and $\Vector{r}$ is constrained to lie on the edge.}

The edge can either be parallel to the face plane or intersecting it.
In both cases, a triangle is defined on a plane that is passing through the centre of mass of the object to minimize the odds of sampling an unstable pose, as done in the face-face pair case.

\textbf{B.1. Intersecting}
If the edge intersects the plane of the face, the intersection point $\Pos{o_0}{o}{o}$ is given by
\begin{align}
    t &= \frac{d - \Normal[\hat]{f}{o} \DotP \Pos{e}{o}{o}}{\Normal[\hat]{f}{o} \DotP \UVec{e}{o}}\\
    \Pos{o_0}{o}{o} &= \Pos{e}{o}{o} + t \UVec{e}{o}.
\end{align}
\rev{The} projections of the centre of mass onto the face and edge are given by
\begin{align}
    \Pos{o_1}{o}{o} &= d\Normal[\hat]{f}{o} + \Pos{c}{o}{o} - \left(\Pos{c}{o}{o}\DotP\Normal[\hat]{f}{o}\right)\Normal[\hat]{f}{o} ,\\
    \Pos{o_2}{o}{o} &= \Pos{e}{o}{o} + \left(\left(\Pos{c}{o}{o} - \Pos{e}{o}{o}\right)\DotP\UVec{e}{o}\right)\UVec{e}{o} ,
\end{align}
respectively, such that a triangle is formed by $\Pos{o_0}{o}{o}$, $\Pos{o_1}{o}{o}$, and $\Pos{o_2}{o}{o}$ in the plane passing through $\Pos{c}{o}{o}$.
\rev{The} method employed in the face-face pair case can \rev{then} be used to find $\Pos{q}{o}{o}$ and $\Pos{r}{o}{o}$.

\textbf{B.2. Parallel}
If the edge is parallel to the face plane, $\Pos{r}{o}{o}$ is set to the projection of the centre of mass onto the edge with
\begin{equation}
    \Pos{r}{o}{o} = \Pos{o_2}{o}{o}.
\end{equation}
\rev{In} the case of the edge not lying in the plane of the face, $\Pos{o_2}{o}{o}$ is moved to the closest point on the face with
\begin{equation}
    \Pos{o_3}{o}{o} = d\Normal[\hat]{f}{o} + \Pos{o_2}{o}{o} - \left(\Pos{o_2}{o}{o}\DotP\Normal[\hat]{f}{o}\right)\Normal[\hat]{f}{o},
\end{equation}
where $\Pos{o_1}{o}{o}$ and $\Pos{o_2}{o}{o}$ are the projections of the centre of mass onto the face and edge respectively as defined previously.
The location of the point on the face is then given by
\begin{equation}
    \Pos{q}{o}{o} = \Pos{o_3}{o}{o} + \frac{\sqrt{(L^2 - \Norm{\Pos{o_3}{o}{o}-\Pos{o_2}{o}{o}}^2)}\left(\Pos{o_1}{o}{o}-\Pos{o_3}{o}{o}\right)}{\Norm{\Pos{o_1}{o}{o}-\Pos{o_3}{o}{o}}}
\end{equation}
such that the projection of the centre of mass onto the face plane lies on the line joining $\Pos{q}{o}{o}$ to $\Pos{r}{o}{o}$ , as desired.

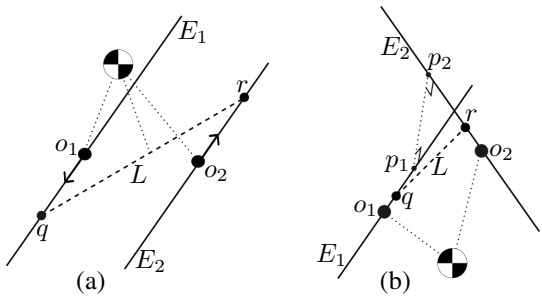
\begin{figure}[b]
    \centering
    \begin{overpic}[width=0.8\linewidth]{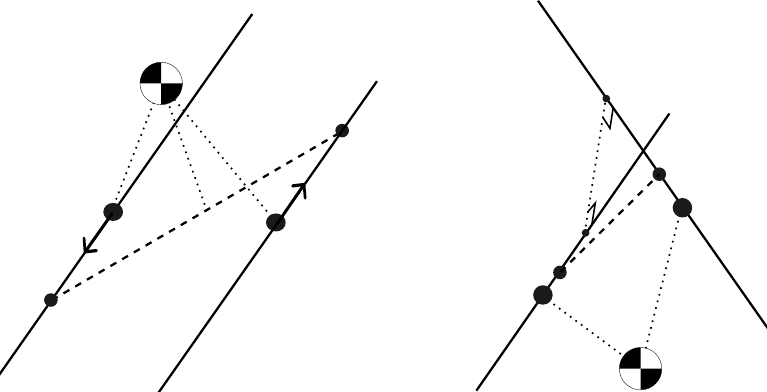}
        \put(13,-2){(a)}
        \put(23,18){$L$}
        \put(32,45){$E_1$}
        \put(24,2){$E_2$}
        \put(9,24){$o_1$}
        \put(37,19){$o_2$}
        \put(5.5,8){$q$}
        \put(43,35){$r$}
        \put(70,-2){(b)}
        \put(79.5,19.5){$L$}
        \put(58,3){$E_1$}
        \put(70,42){$E_2$}
        \put(65,12.5){$o_1$}
        \put(90.5,23){$o_2$}
        \put(74,14){$q$}
        \put(86,29.5){$r$}
        \put(71,21){$p_1$}
        \put(79,40){$p_2$}
    \end{overpic}
    \vspace{0.1cm}
    \caption{Edge-edge pair: (a) parallel, (b) skew.}
    \label{fig:edge_edge_pair}
\end{figure}

\textbf{C. Edge-Edge Pair}
The line vectors of two sampled edges can either be parallel, intersecting, or skew, as shown in \cref{fig:edge_edge_pair}.
For two edge lines defined by their parametric equations
\begin{align}
    E_1(t_1) &: \Pos{q}{o}{o} = \Pos{e_1}{o}{o} + t_1 \UVec{e_1}{o} ,\\
    E_2(t_2) &: \Pos{r}{o}{o} = \Pos{e_2}{o}{o} + t_2 \UVec{e_2}{o} ,
\end{align}
the lines are parallel if $\UVec{e_1}{o} \CrossP \UVec{e_2}{o} = \Vector{0}$.

Unless edges are parallel, there are two points, $\Pos{p_1}{o}{o} = E_1(t_1)$ and $\Pos{p_2}{o}{o} = E_2(t_2)$, that are a minimum distance apart.
These points are given for parameters
\begin{align}
    \label{eqn:lines_closest_points}
    &t_1 = \frac{(\UVec{e_1}{o} \DotP \UVec{e_2}{o})(\UVec{e_2}{o} \DotP \Pos{e_1}{e_2}{o}) - (\UVec{e_1}{o} \DotP \Pos{e_1}{e_2}{o})}{(\UVec{e_1}{o} \DotP \UVec{e_2}{o})(1 - \UVec{e_1}{o} \DotP \UVec{e_2}{o})} ,\\
    &t_2 = \frac{(\UVec{e_2}{o} \DotP \Pos{e_1}{e_2}{o}) - (\UVec{e_1}{o} \DotP \Pos{e_1}{e_2}{o})(\UVec{e_1}{o} \DotP \UVec{e_2}{o})}{(\UVec{e_1}{o} \DotP \UVec{e_2}{o})(1 - \UVec{e_1}{o} \DotP \UVec{e_2}{o})} ,
\end{align}
respectively, where $\Pos{e_1}{e_2}{o} = \Pos{e_1}{o}{o} - \Pos{e_2}{o}{o}$ is the vector joining the origins of the two edges \cite{ericson_real-time_2004}.
If the distance between the two closest points is non-zero, the two edges are skew.
The projection of the centre of mass onto $E_1$ and $E_2$ is given by
\begin{align}
    \label{eqn:com_line_projection_o1}
    \Pos{o_1}{o}{o} &= \Pos{e_1}{o}{o} + \left(\left(\Pos{c}{o}{o} - \Pos{e_1}{o}{o}\right)\DotP\UVec{e_1}{o}\right)\UVec{e_1}{o} ,\\
    \label{eqn:com_line_projection_o2}
    \Pos{o_2}{o}{o} &= \Pos{e_2}{o}{o} + \left(\left(\Pos{c}{o}{o} - \Pos{e_2}{o}{o}\right)\DotP\UVec{e_2}{o}\right)\UVec{e_2}{o} ,
\end{align}
respectively.

\textbf{C.1. Parallel Edge Lines}
\rev{Similar} to previous scenarios, the objective is to select points such that the distance between the object centre of mass and the line joining the two points is minimized, to increase the odds of sampling a stable pose.

For parallel lines, the points $\Pos{q}{o}{o}$ and $\Pos{r}{o}{o}$ are selected by being equally distanced from their respective centre of mass projections in opposite directions with
\begin{align}
    \Pos{q}{o}{o} &= \Pos{o_1}{o}{o} + \frac{1}{2}\sqrt{L^2 - \Norm{\Pos{o_1}{o}{o}-\Pos{o_2}{o}{o}}^2}\UVec{e_1}{o} ,\\
    \Pos{r}{o}{o} &= \Pos{o_2}{o}{o} - \frac{1}{2}\sqrt{L^2 - \Norm{\Pos{o_1}{o}{o}-\Pos{o_2}{o}{o}}^2}\UVec{e_1}{o} ,
\end{align}
such that the two points are distanced by $L$.
If $L < \Norm{\Pos{o_1}{o}{o}-\Pos{o_2}{o}{o}}$, $\Pos{q}{o}{o}$ and $\Pos{r}{o}{o}$ are set to the projections of the centre of mass onto the edges.

\textbf{C.2. Intersecting Edge Lines}
If the edges are intersecting, the intersection point will form a triangle with points $\Pos{q}{o}{o}$ and $\Pos{r}{o}{o}$ on the edges.
However, the two edges are not necessarily part of the same face on non-convex objects.

The intersection point of the two edges $\Pos{o_0}{o}{o}$ can be obtained with \cref{eqn:lines_closest_points} by setting either $t_1$ or $t_2$ in its respective line equation.
The points $\Pos{q}{o}{o}$ and $\Pos{r}{o}{o}$ can be obtained with \cref{eqn:similar_triangle_finding_points} from $\Pos{o_0}{o}{o}$, and the points $\Pos{o_1}{o}{o}$, $\Pos{o_2}{o}{o}$ in \cref{eqn:com_line_projection_o1} and \cref{eqn:com_line_projection_o2}.

\textbf{C.3. Skew Edge Lines}
With skew edges, the closest point on each edge to the other edge, given by \cref{eqn:lines_closest_points}, define a line vector $\Pos{p_1}{p_2}{o} = \Pos{p_1}{o}{o} - \Pos{p_2}{o}{o}$ that is orthogonal to both edges.
Hence, edges lie in parallel planes that are orthogonal to $\Pos{p_1}{p_2}{o}$ and separated by a distance $\Norm{\Pos{p_1}{p_2}{o}}$.
Therefore, the distance between $E_1(t_1)$ and $E_2(t_2)$ is given by 
\begin{equation}
    \Norm{E_2(t_2) - E_1(t_1)} = \sqrt{D^2 + \Norm{\Pos{p_1}{p_2}{o}}^2},
\end{equation}
where $D$ is the distance along the parallel planes, such that the distance between $\Pos{o_1}{o}{o}$ and $\Pos{o_2}{o}{o}$ along the parallel planes is given by $\sqrt{\Norm{\Pos{o_1}{o_2}{o}}^2 - \Norm{\Pos{p_1}{p_2}{o}}^2}$.
Similar to how the position of the contact points were defined in \cref{eqn:similar_triangle_finding_points}, the points $\Pos{q}{o}{o}$ and $\Pos{r}{o}{o}$ are defined with
\begin{align}
    \Pos{q}{o}{o} &= \Pos{p_1}{o}{o} + \frac{\left(\Pos{o_1}{o}{o}-\Pos{p_1}{o}{o}\right)\sqrt{L^2 - \Norm{\Pos{p_1}{p_2}{o}}^2}}{\sqrt{\Norm{\Pos{o_1}{o_2}{o}}^2 - \Norm{\Pos{p_1}{p_2}{o}}^2}} ,\\
    \Pos{r}{o}{o} &= \Pos{p_2}{o}{o} + \frac{\left(\Pos{o_2}{o}{o}-\Pos{p_2}{o}{o}\right)\sqrt{L^2 - \Norm{\Pos{p_1}{p_2}{o}}^2}}{\sqrt{\Norm{\Pos{o_1}{o_2}{o}}^2 - \Norm{\Pos{p_1}{p_2}{o}}^2}} ,
\end{align}
that will produce a solution unless $\Norm{\Pos{p_1}{p_2}{o}} > L$, in which case the edges are too far apart to produce a valid match.
,

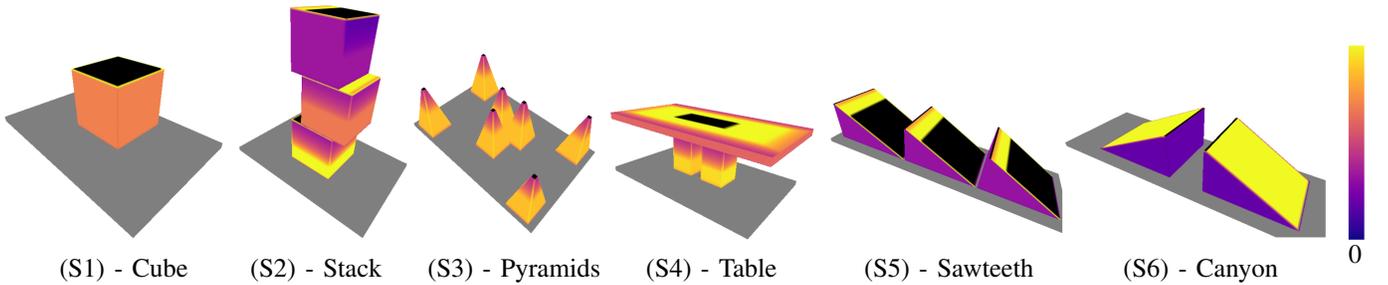
\begin{figure*}[t]
    \centering
    \begin{overpic}
        [width=\linewidth]{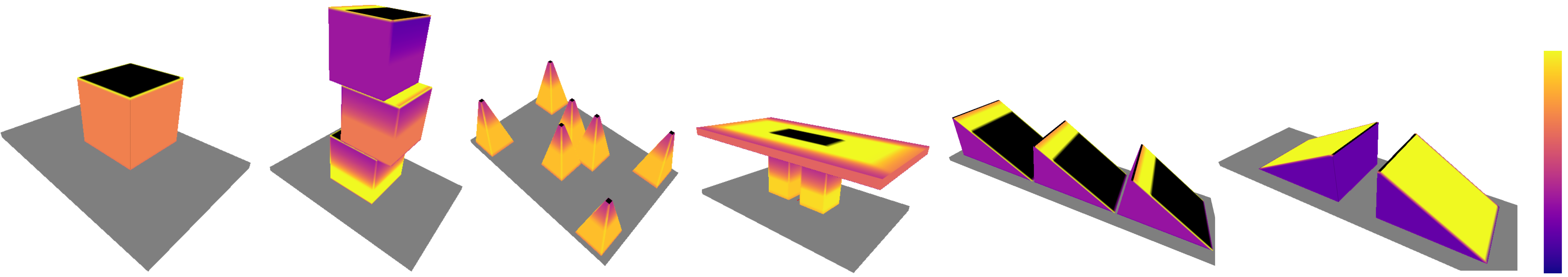}
        \put(4,-2.5){(S1) - Cube}
        \put(18,-2.5){(S2) - Stack}
        \put(31,-2.5){(S3) - Pyramids}
        \put(47,-2.5){(S4) - Table}
        \put(63,-2.5){(S5) - Sawteeth}
        \put(82,-2.5){(S6) - Canyon}
        \put(98.5,-1.5){0}
    \end{overpic}
    \vspace{0.4cm}
    \caption{The six scenes used for the experiments. The surface colour indicates the relative magnitude of the maximal normal force that can be sustained, with a lighter colour indicating a larger force and black indicating an infinite force.}
    \label{fig:scenes}
\end{figure*}

\subsection{Determination of Object Pose}
Let $\Normal[\hat]{e_1}{o}$ be equal to the average of the two face normals that $E_1$ is joining.
By definition, $\Normal[\hat]{e_1}{o}$ will be in the plane orthogonal to $E_1$.
For a placement to be valid, the normal at the scene contact point $\Normal[\hat]{a}{w}$ must also lie in the plane orthogonal to the edge.
Moreover, the angle between $\Normal[\hat]{e_1}{o}$ and $\Normal[\hat]{a}{w}$ must be less than $90^\circ$ for the placement to be non-penetrating.
Unless the two edges are collinear, we have that $\Pos{a}{b}{w}\CrossP\Normal[\hat]{a}{w} \neq 0$ such that there exists a vector
\begin{equation}
    \UVec[\hat]{}{w} = \frac{\Pos{a}{b}{w}\times\Normal[\hat]{a}{w}}{\Norm{\Pos{a}{b}{w}\times\Normal[\hat]{a}{w}}}
\end{equation}
that is orthogonal to the scene face and to $\Pos{a}{b}{w}$, the direction of the line vector connecting the two contact points.
The vector in the object frame that is analogous to $\UVec[\hat]{}{w}$ is given by
\begin{equation}
    \UVec[\hat]{}{o} = -\frac{\Pos{q}{r}{o} \times \Normal[\hat]{e_1}{o}}{\Norm{\Pos{q}{r}{o} \times \Normal[\hat]{e_1}{o}}} ,
\end{equation}
such that
\begin{equation}
    \UVec[\hat]{}{w} = \Rot{o}{w} \UVec[\hat]{}{o}
\end{equation}
relates the two vectors.

Defining \rev{basis vectors}
\begin{align}
    \WVec[\hat]{}{w} &= \frac{\UVec[\hat]{}{w} \CrossP \Pos{a}{b}{w}}{\Norm{\UVec[\hat]{}{w} \CrossP \Pos{a}{b}{w}}} ,\\
    \WVec[\hat]{}{o} &= \frac{\UVec[\hat]{}{o} \CrossP \Pos{q}{r}{o}}{\Norm{\UVec[\hat]{}{o} \CrossP \Pos{q}{r}{o}}} ,
\end{align}
a frame fixed in the scene can be defined with
\begin{equation}
    \CFrame{s} = \bbm \UVec[\hat]{}{w} & \frac{\Pos{a}{b}{w}}{\Norm{\Pos{a}{b}{w}}} & \WVec[\hat]{}{w} \ebm\Transpose .
\end{equation}
\rev{A} frame fixed in the object can be obtained with
\begin{equation}
    \CFrame{o} = \bbm \UVec[\hat]{}{o} & \frac{\Pos{q}{r}{o}}{\Norm{\Pos{q}{r}{o}}} & \WVec[\hat]{}{o} \ebm\Transpose ,
\end{equation}
such that
\begin{equation}
    \Rot{o}{w} = \CFrame{s} \CFrame{o}\Transpose
\end{equation}
is the orientation of the object in the world frame.
With the knowledge of $\Rot{o}{w}$, the position of the object can subsequently be obtained with
\begin{equation}
    \Pos{o}{w}{w} = \Pos{a}{w}{w} - \Rot{o}{w} \Pos{q}{o}{o},
\end{equation}
where $\Pos{a}{w}{w}$ is the position of the first scene contact point.

\subsection{Pose Validation}
Once a pose is determined, it undergoes a series of checks to ensure that the object does not interpenetrate other objects and does not destabilize the assembly.
First, a collision detection algorithm verifies that the object avoids penetrating other assembly components, and detects contact interfaces on which the object is resting.
\rev{To determine the location of contact points, we use the approach from \cite{mattikalli_finding_1996} and select the vertices of the convex hull over the contact interface as contact points, as shown in \cref{fig:toppling_axis}.}
Reaction forces at the contact points are then computed using the QR-based method in \cref{eqn:reaction_forces_solution} and the maximal tension force is compared to a threshold value (e.g., 5 Newtons) to avoid performing the more computationally expensive QP-based method when the placement is clearly unstable.
Finally, the forces at the contact points are computed using \cref{eqn:optim_problem}, the equilibrium of the assembly is verified, and the static robustness of the assembly is updated.

\section{Simulation Experiments}
\label{sec:experiments}
\begin{table*}
    \centering
    \caption{Average placement time in seconds, robustness to external wrench (Rob.) in Newtons, minimum static robustness (Min. SR) in Newtons, and volume of the assembly in cubic meters for each algorithm across our six scenes. Entries with --- indicate that no valid placement was found in the experiments}
    \label{tab:results}
    \begin{tabular}[t]{cc|c|c|c|c|c|c}
        \toprule
        & & \textbf{Chance} & \textbf{Sim-Random} & \textbf{Sim-Above} & \textbf{HM} & \textbf{Ours-Uniform} & \textbf{Ours-SR}\\
        \midrule
        \multirow{4}{*}{\rotatebox{90}{Cube}} & Time &2.85 &11.23 &0.79 &0.88 &3.60 &\textbf{0.38}\\
        & Rob. &--- &1.72 &1.99 &\textbf{3.21} &1.64 &1.81\\
        & Min. SR &--- &2.29 &2.64 &\textbf{4.44} &2.21 &2.47\\
        & Volume &--- &30.0 &28.4 &\textbf{17.2} &30.1 &28.0\\
        \midrule
        \multirow{4}{*}{\rotatebox{90}{Stack}} & Time &4.88 &54.5 &7.76 &\textbf{0.56} &120 &1.51\\
        & Rob. &--- &0.48 &0.41 &\textbf{0.80} &0.53 &0.56\\
        & Min. SR &--- &1.76 &1.88 &\textbf{2.40} &2.10 &2.00\\
        & Volume &--- &70.6 &66.7 &\textbf{47.3} &58.0 &61.1\\
        \midrule
        \multirow{4}{*}{\rotatebox{90}{Table}} & Time &5.13 &20.8 &6.59 &0.83 &13.8 &\textbf{0.40}\\
        & Rob. &--- &1.76 &1.85 &1.11 &1.86 &\textbf{1.95}\\
        & Min. SR &--- &2.10 &2.10 &2.10 &1.65 &2.10\\
        & Volume &--- &105 &103 &109 &\textbf{88.3} &101\\
        \midrule
        \multirow{4}{*}{\rotatebox{90}{Pyramids}} & Time &6.47 &97.25 &41.74 &\textbf{1.10} &18.62 &1.71\\
        & Rob. &--- &0.78 &0.75 &\textbf{0.85} &0.49 &0.40\\
        & Min. SR &--- &\textbf{1.69} &1.10 &1.54 &0.41 &0.42\\
        & Volume &--- &80.1 &\textbf{72.4} &128 &96.8 &135\\
        \midrule
        \multirow{4}{*}{\rotatebox{90}{Sawteeth}} & Time &4.30 &9.43 &8.74 &30.0 &10.4 &\textbf{2.86}\\
        & Rob. &--- &0.71 &0.64 &--- &0.91 &0.77\\
        & Min. SR &--- &1.91 &1.80 &--- &1.58 &1.73\\
        & Volume &--- &\textbf{136} &139 &--- &161 &159\\
        \midrule
        \multirow{4}{*}{\rotatebox{90}{Canyon}} & Time &4.02 &55.2 &45.7 &31.6 &9.85 &\textbf{1.30}\\
        & Rob. &--- &0.85 &0.88 &--- &1.40 &\textbf{1.68}\\
        & Min. SR &--- &0.57 &0.29 &--- &0.52 &\textbf{0.72}\\
        & Volume &--- &109 &\textbf{108} &--- &130 &128\\
        \midrule
        \multicolumn{2}{c|}{\textbf{Average Time}} &4.61 &41.40 &18.55 &10.83 &29.38 &\textbf{1.36}\\
        \multicolumn{2}{c|}{\textbf{Success Rate}} &0\% &93.3\% &99.7\% &66.7\% &95.3\% &\textbf{100\%}\\
        \bottomrule
    \end{tabular}
\end{table*}

\rev{We perform a series of simulation experiments} across six different scenes, shown in \cref{fig:scenes}, where the task is to place a cube such that it stably rests on other objects.
Each scene is designed to challenge the algorithms in different ways.
\rev{The} \textit{Stack} scene has many points on vertical surfaces, \textit{Pyramids} has few robust contact points, \textit{Table} is prone to toppling, \textit{Sawteeth} has no horizontal surfaces, and \textit{Canyon} only affords placements resting on both objects.
The benchmarked algorithms and evaluation criteria are described in \cref{sec:benchmarked_algorithms} and \cref{sec:evaluation_criteria} respectively, and the results are reported in \cref{tab:results}.
For all sample-based planners, fifty consecutive experiments are performed for each scene with a maximum of 500 attempts per experiment.
\rev{Algorithms are evaluated} in terms of planning time, assembly robustness, packing density, and success rate.

\subsection{Benchmarked Algorithms}
\label{sec:benchmarked_algorithms}
\subsubsection{Planner Using Static Robustness (Ours-SR)}
We compare to other methods our proposed algorithm defined in \cref{sec:placement_planning_from_static_robustness} and using \cref{eqn:prob_dist_with_robustness} to sample points, with the initial probability of sampling the most robust point set to $10\%$ and the exponential decay rate $\lambda$ set to $0.99$.
In all experiments involving contact point sampling, if multiple pairs of features can be matched to the sample points, a single random pair is selected.

\subsubsection{Planner With Uniform Sampling (Ours-Uniform)}
In essence, our algorithm, defined in \cref{sec:placement_planning_from_static_robustness}, uses the proposed static robustness heuristic to increase the odds of sampling promising contact points in the scene.
Insights about the effectiveness of the proposed algorithm can be obtained by comparing it with a variation for which the probability of sampling a point is uniform.
If there is a significant benefit to using the static robustness heuristic, the proposed algorithm should perform better when the probability distribution in \cref{eqn:prob_dist_with_robustness} is used, and worst when it is uniform.

\subsubsection{Dynamics Simulation From Above (Sim-Above)}
A dynamics simulator is also used to provide a standard for comparison.
Object placement from simulations is performed in two phases during which the restitution coefficient is set to zero, and the dynamics are heavily damped to reduce any bouncing phenomena.
In the first phase, the object is dropped in a random orientation with its centre of mass at a random position slightly above the non-fixed objects in the scene.
\rev{The} gravitational acceleration is set to 0.1 $\textrm{m}/\textrm{s}^2$ and the downward velocity of the object is set to 0.01 $\textrm{m}/\textrm{s}$ to reduce the kinetic energy of the object when it collides with the scene.
The simulation is run until the object comes to rest and is aborted if the object falls off the scene floor.
In the second phase, gravity is increased to 9.81 $\textrm{m}/\textrm{s}^2$ and the simulation is ran for 100 iterations to ensure that the placement is stable under standard gravitational acceleration.
If the placed object rests on at least one other object, and no object has moved significantly, the placement is considered to be successful.

\subsubsection{Dynamics Simulation From A Random Pose (Sim-Random)}
Performing simulations from initial poses above the scene objects will not allow the planner to find placement poses below other objects.
A more general approach is to perform simulations from random poses, where the pose of the object is initialized randomly within the vicinity of the scene objects.
When sampling positions, the extent of the scene was defined from the bounding box of the objects in the scene and points around the scene, up to a distance equal to the radius of the sphere circumscribing the object, were considered.
Otherwise, all parameters are kept the same as in the \textit{Sim-above} algorithm.

\subsubsection{Heightmap Minimization (HM)}
The algorithm defined in \cite{wang_dense_2021} is implemented and used as a benchmark. 
A summary of the algorithm is provided below.
Planning with the \textit{HM} algorithm is performed in four steps: (i) candidate pose generation, (ii) scoring, (iii) stability evaluation, and potentially (iv) refined search.
The first step uses \cite{goldberg_part_1999} to generate four object orientations that are likely stable on planar surfaces, and perturb the orientations with rotations about the vertical axis.
The scene is voxelized and a set of candiate poses is generated by finding the lowest point in the voxel grid for each object orientation.
The second step scores the candidate poses with a heightmap minimization heuristic proposed in \cite{wang_dense_2021} that encourages dense and stable placements. 
The third step iterates over the highest scoring poses and evaluates their stability by solving \cref{eqn:optim_problem} and returning the first stable pose found.
Finally, in the event that no stable pose is found, the search is refined by generating a larger set of candidate poses through rotation perturbations about orthogonal axes in the horizontal plane.
Our implementation of \textit{HM} uses the same parameters and the same voxel resolution as in \cite{wang_dense_2021}.

\subsubsection{Random Pose (Chance)}
To provide a baseline for comparison, a random pose within the vicinity of the scene objects is selected for the object, and the stability of the placement is evaluated.
Although this method is expected to be the slowest, it should be capable of finding stable placements if enough iterations are performed (although a maximum of 500 attempts are made in the experiments).

\subsection{Evaluation Criteria}
\label{sec:evaluation_criteria}
Five criteria are used to evaluate the performance of the algorithms.

\subsubsection{Placement Time}
For each set of experiments, the average time required to find a collision-free stable placement pose is measured.
Since some algorithms may require more iterations but less compute time per iteration, all algorithms are compared in terms of the average compute time per valid placement.  

\subsubsection{Assembly Robustness}
\rev{The} objective of inertia-aware object placement planning is to find placements that maximize the force required to displace any object in the assembly.
\rev{Hence,} the minimal magnitude of the external wrench that would destabilize the assembly is computed according to \cite{maeda_new_2009, chen_planning_2021}, where the wrench directions are defined as the orthogonal axes in the 6D wrench space.

\subsubsection{Minimum Static Robustness}
Our static robustness heuristic defined in \cref{sec:static_robustness_assessment} is computed for points on all objects in the scene following a valid placement.
The minimum static robustness, representing the smallest force that can be applied to the surface of an object such that the assembly is destabilized, is computed for each experiment.

\subsubsection{Packing Density}
Since it is often desired to place objects such that the assembly is dense, the packing density of the assembly is computed as the volume of the oriented bounding box of the assembly following the placement.

\subsubsection{Success Rate}
The proportion of experiments in which a valid placement was found is reported for each algorithm.

\section{Real Robot Experiments}
\label{sec:real_robot_experiments}
\begin{figure}
    \centering
    \begin{overpic}[width=1\linewidth]{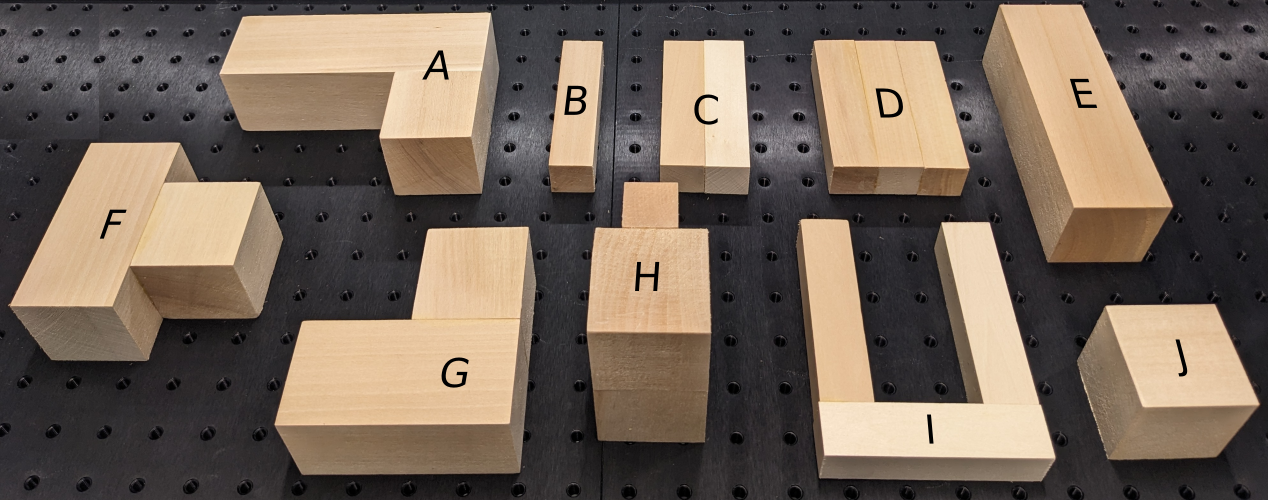}
    \end{overpic}
    \caption{The set of 10 basswood objects used for the real robot experiments.}
    \label{fig:wooden_blocks}
\end{figure}

A practical placement planner should produce stable assemblies under the presence of uncertainty in the pose, shape, and inertial parameters of the objects.
To evaluate the effectiveness of the proposed algorithm in such a scenario, we performed experiments with a Franka Research 3 robot arm equipped with a Robotiq 2F-85 gripper.
A set of 10 non-convex basswood objects, pictured in \cref{fig:wooden_blocks}, is used for this experiment, where each object is built from manually-glued cuboids resulting in shapes that are slightly different from the models used by the planner.
The mass density of the objects is assumed to be 415 $\textrm{kg}/\textrm{m}^3$, which is expected to differ from the actual value, introducing errors in the inertial models of the objects.
\rev{In our experiments, the friction coefficient was assumed to be $0.5$ at interfaces between our basswood objects, which is expected to be an optimistic value.}
\rev{In practical applications where stability is critical, the friction coefficient can be assumed to be lower than its nominal value such that the planner acts more conservatively.}
Uncertainty in the initial pose of the objects, slippage, and Cartesian trajectory errors are expected to create a discrepancy between the planned and executed placements, potentially resulting in unstable assemblies. 

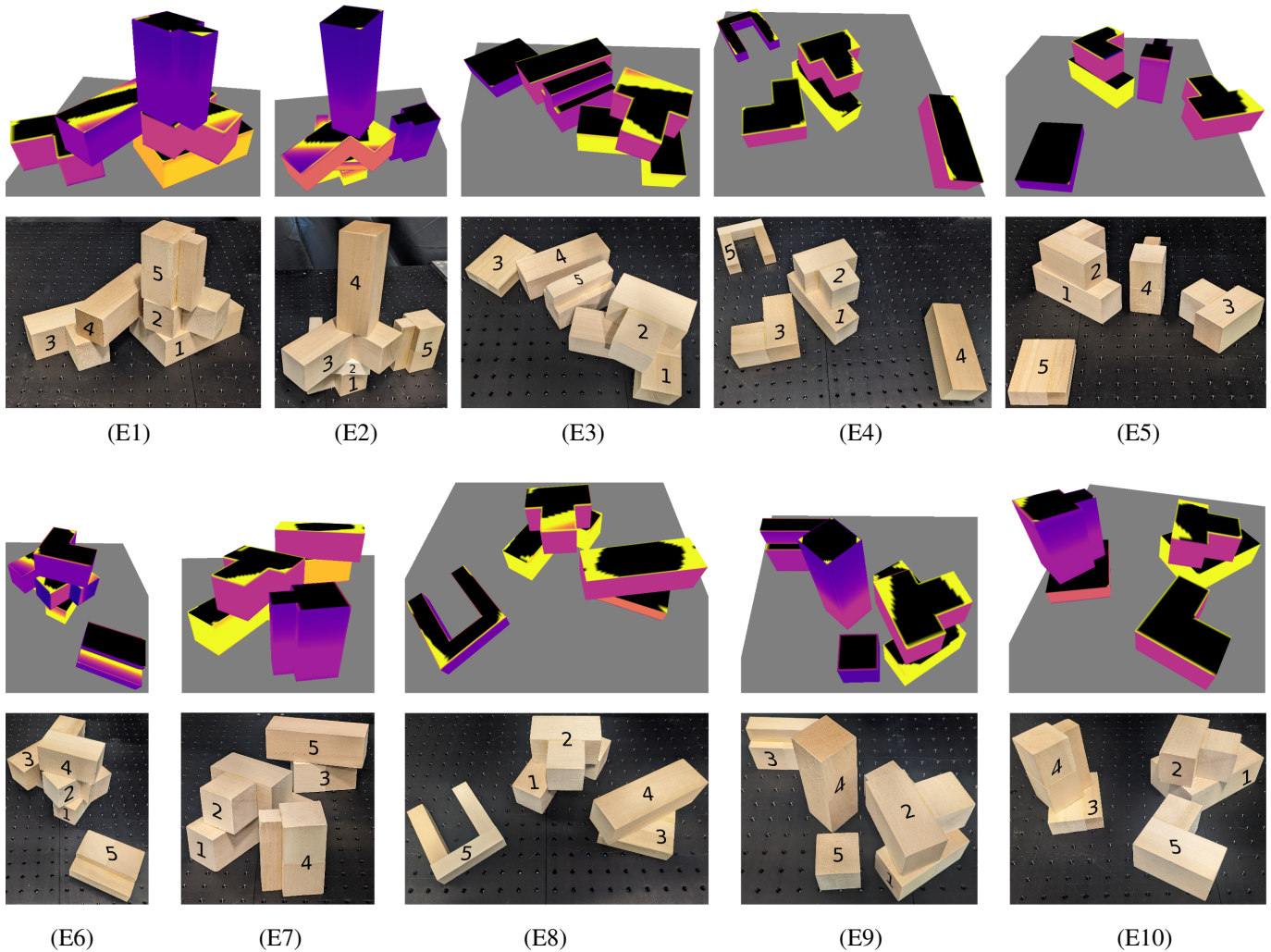
\begin{figure*}[h]
    \centering
    \begin{overpic}
        [width=\linewidth]{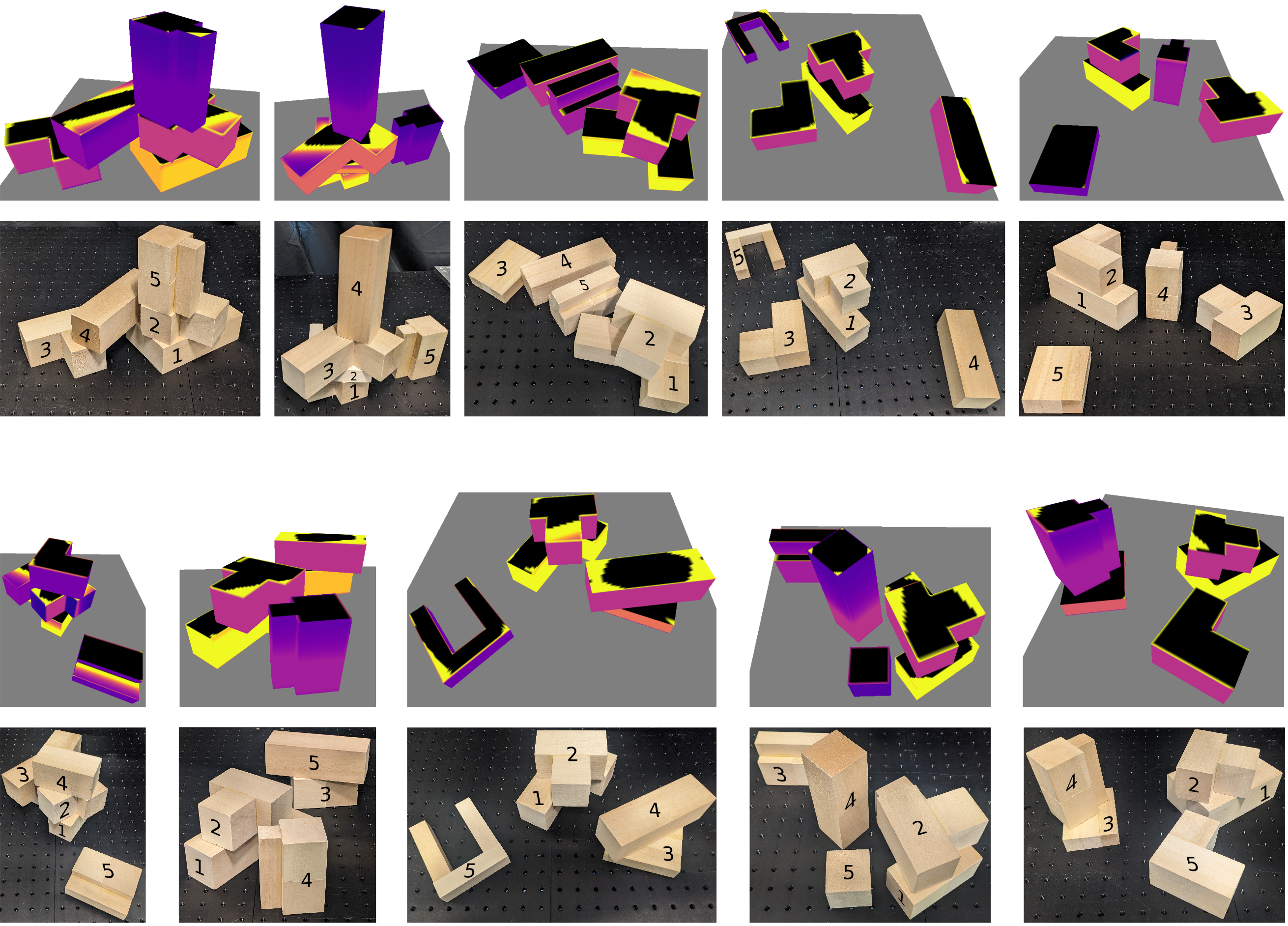}
        \put(8,37){(E1)}
        \put(26,37){(E2)}
        \put(44,37){(E3)}
        \put(66,37){(E4)}
        \put(88,37){(E5)}
        \put(3.5,-3){(E6)}
        \put(20,-3){(E7)}
        \put(41,-3){(E8)}
        \put(66,-3){(E9)}
        \put(88,-3){(E10)}
    \end{overpic}
    \vspace{3mm}
    \caption{The ten scenes from the real robot experiments with the SR map in the top row and the final assembly in the bottom row. The order in which objects are placed is indicated by the numbers on the objects.}
    \label{fig:real_robot_scenes}
\end{figure*}

A total of 50 placements from 10 experiments pictured in \cref{fig:real_robot_scenes} (E1 to E10) are performed with the robot. 
Each experiment consists of iteratively placing objects in the scene until five objects are placed.
For each placement, the proposed planner is ran 10 times, each time executing at most 20 iterations with an object selected randomly without replacement and with odds proportional to its mass (i.e., heavier objects are more likely to be placed first).
For these experiments, sampling on the fixed support is allowed and \cref{eqn:prob_dist_with_robustness_fixed} is used with $\gamma =$ 0.5 and with other parameters set to the same values as in the simulation experiments.
If more than one placement results in the same minimum SR, the one with the largest median SR is selected in E1--E5, and the one with the smallest assembly volume is selected in E6--E10. 

With the placement pose defined, a feasible grasp as far as possible to the contact points is determined. %
For each placement, the RRTConnect \cite{kuffner_rrt_2000} motion planner was run twice, each time for at most 10 seconds, to find collision-free trajectories to pick up and place the object.

The success of any placement depends on five cascading steps: (i) finding a feasible placement pose, (ii) grasping the object, (iii) planning a collision-free motion trajectory, (iv) executing the placement without slippage or collisions, and (v) keeping the assembly stable.
During our experiments, human observation was used to determine the outcome of each step and any issue that arose (e.g. slip, contact with the environment, toppled assembly) was noted.
The human observer manually carried out any uncompleted placement by approximately placing the object in its goal pose such that the following placements could be subsequently performed by the robot.
The success rate of each step relative to the number of placements that reached the step is reported in \cref{tab:real_robot_results}. 

\begin{table}
    \centering
    \caption{Success rate of each step in the real robot experiments with 50 placements, computed relative to the number of successful attempts from the previous step.}
    \label{tab:real_robot_results}
    \begin{tabular}{l|c}
        \toprule
        Step & Success Rate\\
        \midrule
        Placement Planning  & 50/50 = 100\%\\
        Grasping            & 41/50 = 82\%\\
        Motion Planning     & 37/41 = 90.2\%\\
        Placement Execution & 33/37 = 89.2\%\\
        Stable Assembly     & 33/33 = 100\%\\
        \bottomrule
    \end{tabular}
\end{table}

\section{Discussion}
\label{sec:discussion}

The performance of the proposed algorithm in comparison with other algorithms is analyzed in \cref{sec:performance_analysis}.
\rev{Observations} made during the real robot experiments are outlined in \cref{sec:real_robot_observations} and the computational complexity of the proposed algorithm is analyzed in \cref{sec:computational_complexity}.

\subsection{Performance Analysis}
\label{sec:performance_analysis}
In our experiments, \textit{Chance} never found a valid placement in the 3,000 attempts made, indicating that the placement planning problem is not trivial.

According to \cref{tab:results}, the proposed algorithm using static robustness was about 20 times faster to find a stable placement when compared to the same algorithm without static robustness and about eight times faster than the \textit{HM} algorithm.
The simulator also took much longer than the proposed algorithm to find a valid solution, on average taking about 13.6 times longer with \textit{Sim-Above} and 30 times longer with \textit{Sim-Random}.
\rev{This poor performance} is expected, given that the \rev{simulator needlessly} solves a much more complex problem (i.e., coupled kinematics and dynamics) \cite{kaufman_coupled_2009}.

Compared to other algorithms, \textit{HM} can find placements that are more robust to external wrenches.
However, \textit{HM} found valid placements in only 66.7\% of the experiments, while the other algorithms (except \textit{Chance}) found valid placements in 93\% to 100\% of the experiments.
Even though \textit{HM} optimizes for packing density while \textit{Ours-SR} does not, both algorithms produce assemblies with similar volumes.
In contrast, the other algorithms can produce better packed assemblies by having the object be supported by the floor (e.g. under the table).

Algorithms that sample uniformly struggle in large scenes that afford only few placements, as seen in the \textit{Stack} scene where the time required by \textit{Sim-Random} and \textit{Ours-Uniform} is significantly higher than for the other algorithms.
In the \textit{Table} scene, the robustness to external wrench is significantly lower for \textit{HM} since it selects the first stable pose found, which is not the one in the centre of the table.
However, in \textit{Pyramids}, \textit{Ours-SR} finds placements with relatively low robustness compared to \textit{HM} since the former does not centre the cube on the three pyramids.
By virtue of the three pyramids being identical, \textit{HM} produces better centred placements, resulting in a more robust assembly.
In the \textit{Sawteeth} and \textit{Canyon} scenes, \textit{HM} cannot find any valid placements since it iterates over the $N$ highest scoring placements, which are all unstable in these scenes.
This issue could be alleviated by performing stability checks at the candidate pose generation stage, at the cost of a prohibitive increase in computation time due to \textit{HM}'s iteration over all discretized positions and orientations of the object.

\subsection{Observations In Real Robot Experiments}
\label{sec:real_robot_observations}
Four objects in the set (B, C, D, and I) are relatively long but thin, making them quite weak when placed upright and quite robust when placed lying down.
Two of these objects (D and I) were frequently selected by the planner in the experiments (i.e., in E2--E5, E8, E10), and a lying down pose was always selected for them, which is expected from a SR point of view.
However, the objects being thin, the gripper could not grasp them when they were lying down, ultimately resulting in a grasp planning failure.
Similarly, although the pose selected for object H in E3 and object J in E9 are collision-free and make sense from a SR point of view, placing them would have caused the gripper to collide with other objects in the scene.
This issue suggests that the planning of the grasping and release phases should be better integrated with placement planning.
\rev{A} rejection sampling step as done in \cite{chen_planning_2021, wang_dense_2021} could be performed to ensure that the object can be grasped and placed in the desired pose.
\rev{Also, scene and object points that generated a grasp planning failure could have their likelihood of being selected again reduced in subsequent iterations with a slight modification to \cref{eqn:prob_dist_with_robustness}.}
\rev{In two occasions}, the motion planner could not find a path to the placement pose in the allocated time, highlighting that planning for placements that can be easily executed is an avenue for future work.

A supplemental video, from which \cref{fig:frontpage} is taken, is provided\footnote{\rev{At the following URL: http://tiny.cc/stableplacement}} to illustrate the real robot experiments. 
\rev{The accompanying video} highlights (i) a case where the robot successfully places objects in a scene and (ii) another where the robot topples the assembly when placing the last object.
In the first case, when the robot placed the last object whose static robustness is deemed to be very low, a slight pose error almost caused the neighbouring object to topple, effectively validating the computed static robustness.
In the second case, the robot created a contact with a neighbouring object when placing the last object due to an erroneous motion trajectory.
\rev{This contact} exerted force in a direction in which an infinite amount of force could be sustained according to the computed static robustness.
As the robot applied force, the whole structure collapsed, outlining the importance of using static robustness not only to define the placement pose, but also the approach direction of the robot when placing an object.

\subsection{Computational Complexity}
\label{sec:computational_complexity}
As expected, in terms of compute time, the contact resolution and constrained optimization steps are the most demanding, taking together more than 80\% of the total compute time according to \cref{tab:compute_time}.
The QR-based reaction force solver being significantly faster than the QP-based solver confirms that the latter should be used only to verify placements that are likely to be successful.
Per iteration, the overhead of computing the static robustness heuristic accounts for about 10\% of the compute time, which is largely outweighed by the time saved with the reduction in the number of iterations required to find a solution.

\begin{table}
    \centering
    \caption{Average compute time per iteration for each step.}
    \label{tab:compute_time}
    \begin{tabular}[t]{c|c|c}
        \toprule
        & Compute Time (ms) & Proportion\\
        \midrule
        QR-based Reaction Forces     & 9.8   & 6.8\%\\
        QP-based Reaction Forces     & 82.3   & 57.3\%\\
        Static Robustness   & 15.0   & 10.4\%\\
        Pose Definition     & 0.6  & 0.4\%\\
        Contact Resolution & 35.9    & 25.0\%\\
        \midrule
        Total & 144 & 100\%\\
        \bottomrule
    \end{tabular}
\end{table}

\begin{figure*}
    \centering
    \begin{subfigure}{0.30\linewidth}
        \centering
        \includegraphics[width=\linewidth]{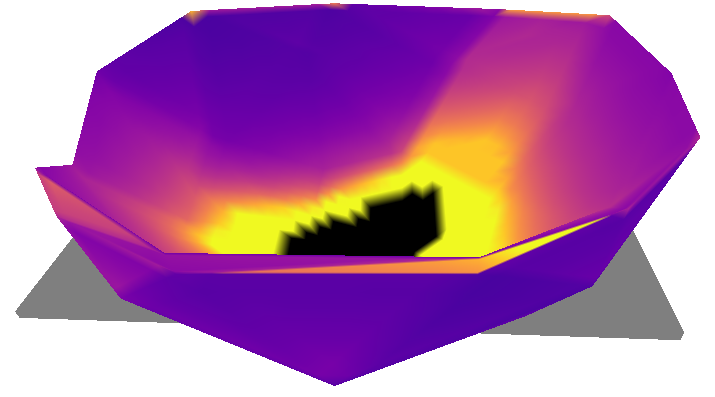}
        \caption{50 vertices}
        \label{fig:bowl_100}
    \end{subfigure}%
    \begin{subfigure}{0.30\linewidth}
        \centering
        \includegraphics[width=\linewidth]{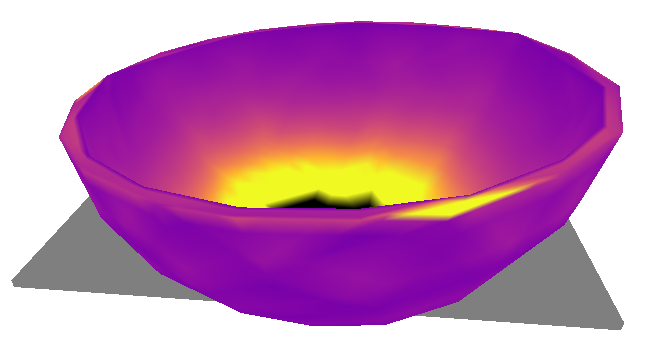}
        \caption{150 vertices}
        \label{fig:bowl_300}
    \end{subfigure}%
    \begin{subfigure}{0.30\linewidth}
        \centering
        \includegraphics[width=\linewidth]{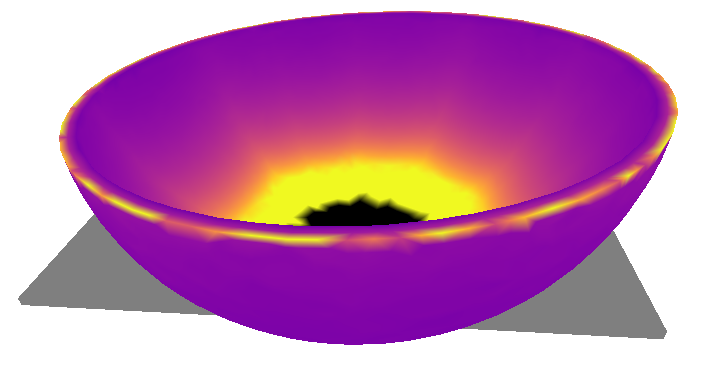}
        \caption{1500 vertices}
        \label{fig:bowl_3000}
    \end{subfigure}
    \caption{The bowl, with various vertex number and overlaid SR map, used to study the computational complexity of the proposed algorithm with curved objects approximated by triangle meshes.}
    \label{fig:bowl_meshes}
\end{figure*}

\begin{figure}
    \centering
    \includegraphics[trim=20 65 10 10, clip, width=1\linewidth]{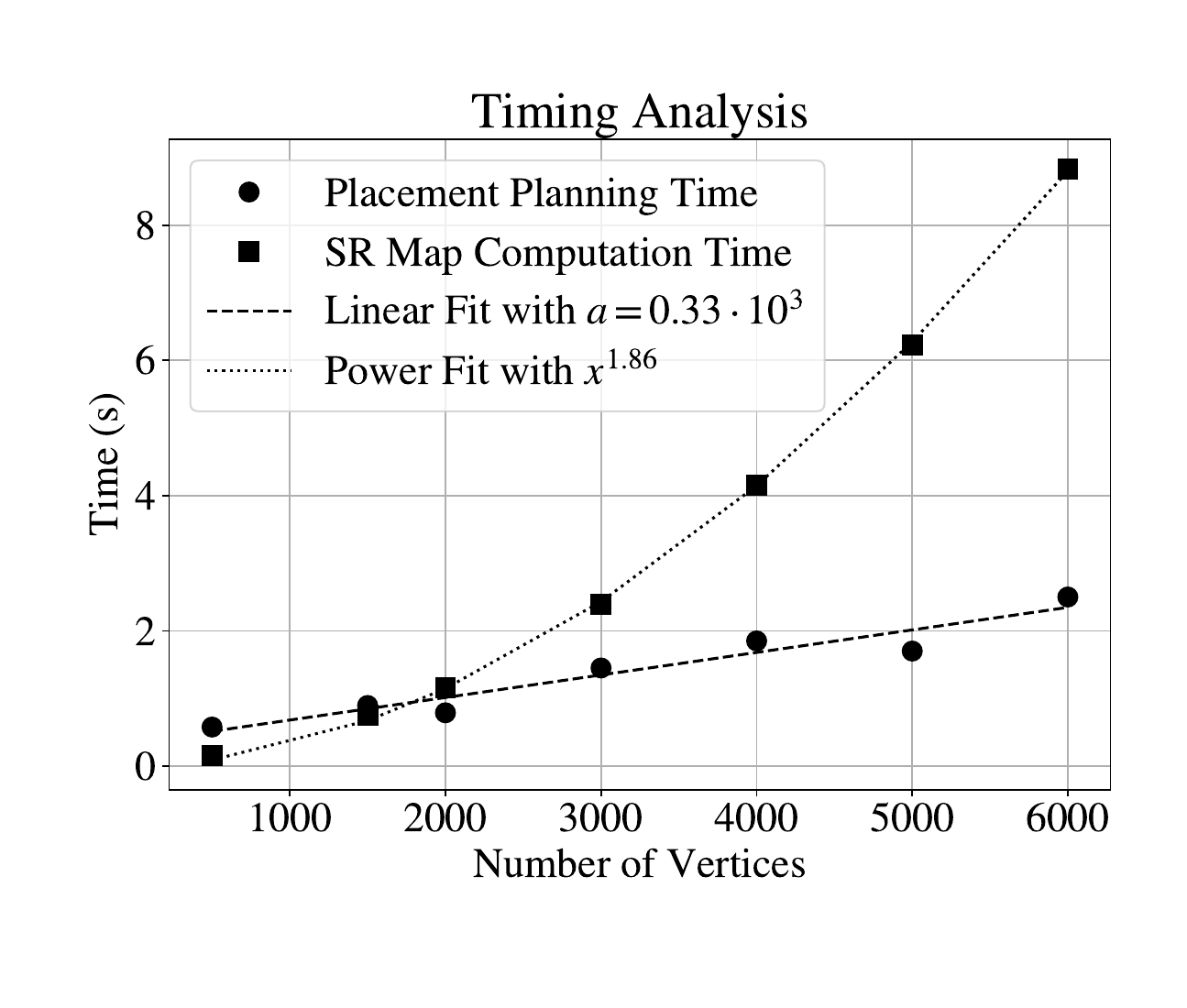}
    \caption{The time required to compute the SR map grows less than quadratically with the number of vertices in the object mesh, and the time required to find a stable placement grows linearly with the number of vertices.}
    \label{fig:complexity}
\end{figure}

\rev{By design, the computation time of the SR map grows linearly with the number of objects in the scene.}
\rev{This is due to each object being considered in isolation in the computation of the SR map.}
\rev{Hence, in a scene comprising a large number of objects, the proposed heuristic should enable our placement planner to focus on subsets of the scene that are more likely to offer a strong support.}

Objects with curved shapes can be challenging for the proposed algorithm, which relies on planar interfaces.
The shape of a curved object, like the bowl shown in \cref{fig:bowl_meshes}, can be approximated by a triangle mesh, with a lower number of vertices producing a coarser approximation that might result in erroneous placement planning.
While a higher number of vertices usually reduces the modeling error, it also increases the computational cost of the algorithm and ultimately slows down the planning process.

Placement planning experiments on a hemispherical shell with flattened end (i.e., a bowl), were performed to study the computational complexity of the proposed algorithm.
Triangle meshes with various numbers of vertices were obtained through quadric decimation \cite{garland_surface_1997} and used by the proposed algorithm to find stable placements for a cube.
Ten experiments were performed for each mesh, and the average time required to compute the SR map and find a stable placement was recorded.
Results from these experiments are shown in \cref{fig:complexity}, where the time required to compute the SR map grows approximately following
\begin{equation}
    t_{\text{SR}} = 8.53\cdot 10^{-7} \cdot v^{1.856} ,
\end{equation}
and the time required to find a stable placement grows approximately following
\begin{equation}
    t_{\text{plan}} = 0.346 + 0.333\cdot 10^{3} v ,
\end{equation}
where $v$ is the number of vertices in the object mesh.
Hence, the time required to compute the SR map grows less than quadratically with the number of vertices in the object mesh, and the time required to find a stable placement grows linearly with the number of vertices.
\rev{However, as can be noticed in \cref{fig:bowl_meshes}, there is a diminishing return to increasing the resolution of the SR map beyond a certain point with only a subtle difference between \hyperref[fig:bowl_meshes]{\cref*{fig:bowl_meshes}b} and \hyperref[fig:bowl_meshes]{\cref*{fig:bowl_meshes}c}, with the latter computed at a $10\times$ greater resolution.}
Consequently, an adequate mesh resolution should be selected to balance the trade-off between modeling error and computational complexity.
\rev{Although the execution time of the proposed algorithm does not grow prohibitively with the number of vertices, the processing time can be limited by using a coarser resolution for SR computations and an interpolation scheme to produce a finer map.}

\section{Conclusion}
\label{sec:conclusion}
This work defines an inertia-aware object placement planner that can scale to scenes with many objects with no assumption made on object convexity, homogeneous density, or shape.
Our algorithm makes use of the object inertial parameters at every step of the process to increase the odds of sampling a stable placement pose and to preserve the assembly's robustness to external perturbations.
We show that the use of our \textit{static robustness} heuristic greatly increases the odds of sampling a stable placement pose, with a light computational overhead, making the proposed algorithm much faster than other methods.
Future work will focus on improving the efficiency of the algorithm by performing a local optimization after having defined a placement pose such that an almost stable pose can be adjusted to be stable.
Also, a learning-based approach to static robustness inference will be investigated to improve compute time while keeping the generalizability of the method.

\bibliographystyle{ieeetr}
\bibliography{extracted}

\begin{IEEEbiography}[{\includegraphics[width=1in,height=1.25in,trim={8mm 0mm 8mm 0},clip,keepaspectratio]{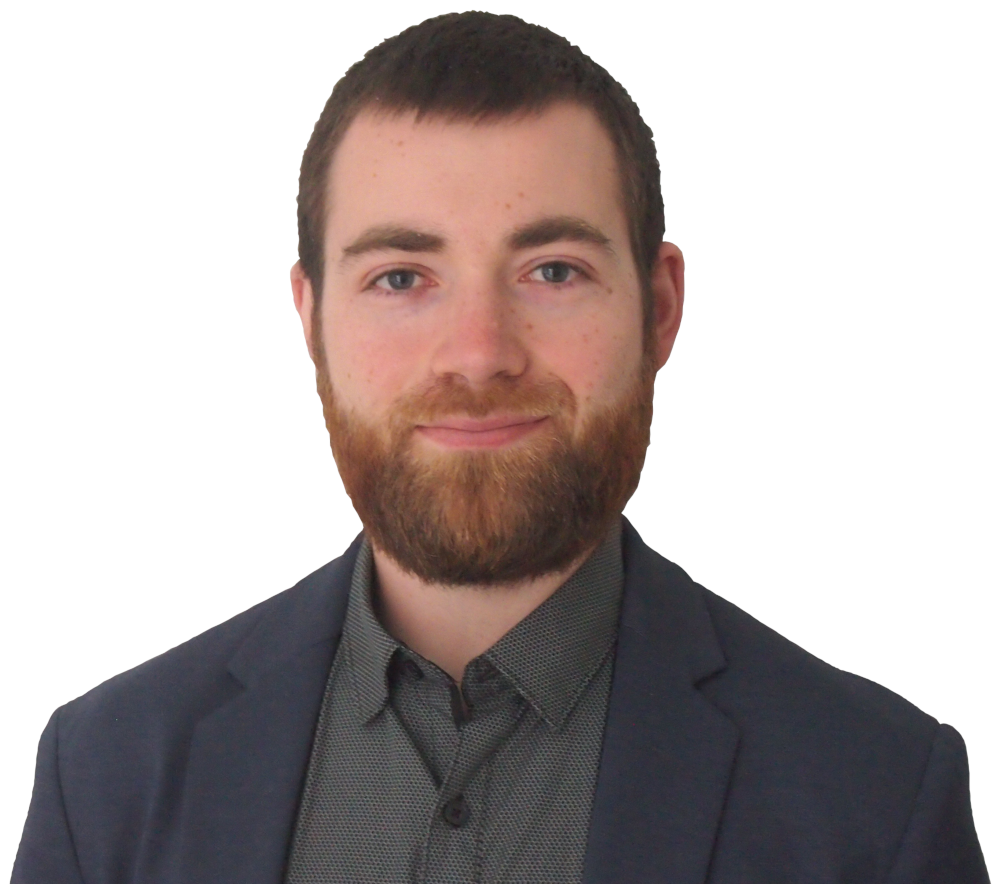}\vspace{-5mm}}]{Philippe Nadeau} received his B.Eng degree, with honours, in automated manufacturing engineering from École de Technologie Supérieure, Montréal, Canada.
He is currently a Ph.D. candidate in the STARS laboratory at the University of Toronto Institute for Aerospace Studies where he works on designing novel algorithms to enhance collaborative robots autonomy in object handling tasks.
Philippe received various awards for his research work, including Canada's Graduate Scholarships, Québec's Master's Research Scholarship, and Vector Institute's Scholarship in Artificial Intelligence.
His research interests lie at the intersection of perception and planning with an emphasis on fast and general algorithms deployed on real robots in human-centric environments.
\end{IEEEbiography}

\begin{IEEEbiography}[{\includegraphics[width=1in,height=1.25in,clip,keepaspectratio]{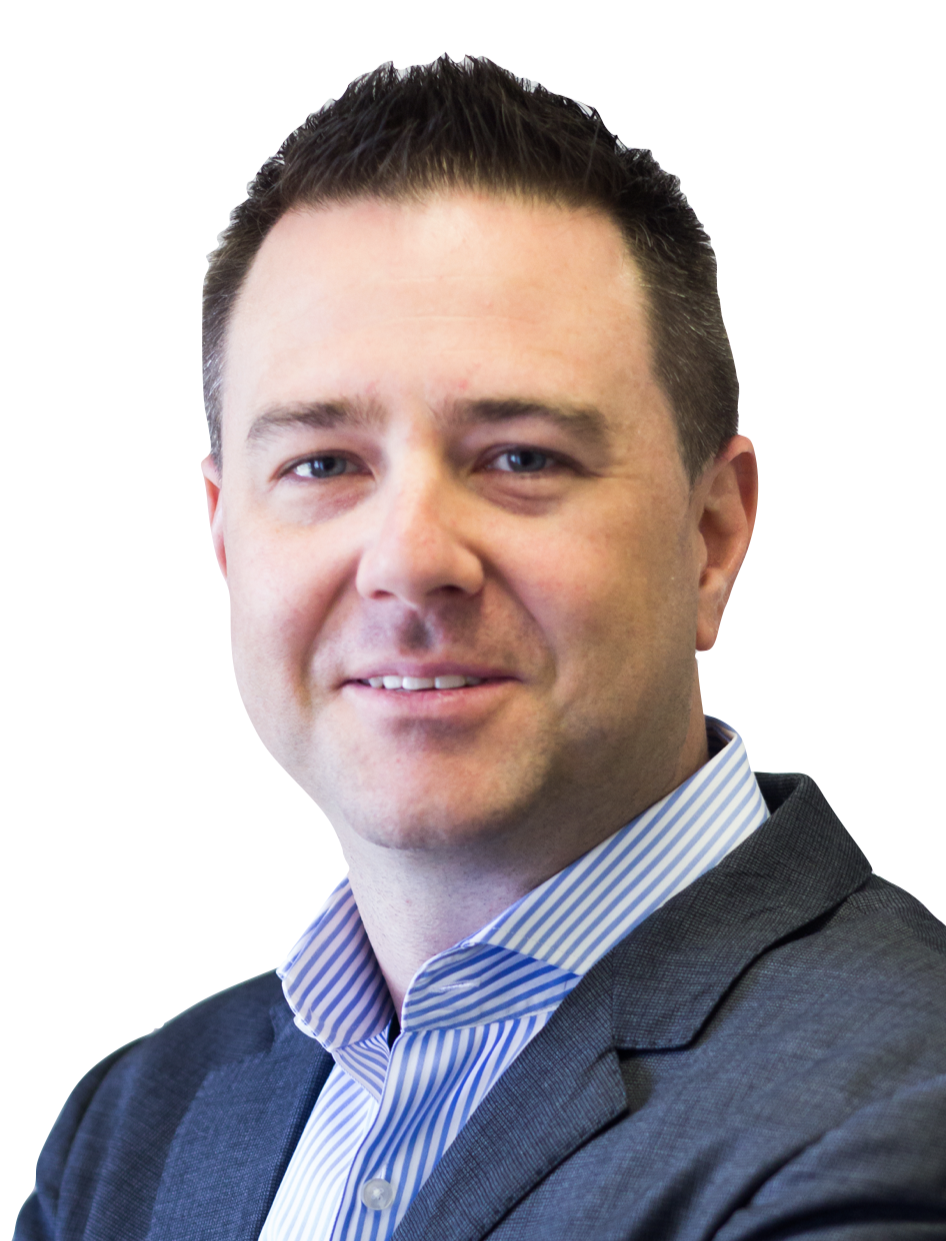}}]{Jonathan Kelly} received his Ph.D.\ degree from the University of Southern California, Los Angeles, USA, in 2011. From 2011 to 2013 he was a postdoctoral fellow in the Computer Science and Artificial Intelligence Laboratory at the Massachusetts Institute of Technology, Cambridge, USA. He is currently an associate professor and director of the Space and Terrestrial Autonomous Robotic Systems (STARS) Laboratory, University of Toronto Institute for Aerospace Studies, Toronto, Canada. Prof. Kelly holds the Canada Research Chair in Collaborative Robotics. His research interests include perception, planning, and learning for interactive robotic systems.
\end{IEEEbiography}

\vfill

\end{document}